\RequirePackage{fix-cm}
\documentclass[11pt]{article}

\usepackage[preprint]{acl}
\usepackage{times}
\usepackage{latexsym}
\usepackage[T1]{fontenc}
\ifx\XeTeXversion\undefined\usepackage[utf8]{inputenc}\fi
\usepackage{microtype}
\usepackage{inconsolata}

\usepackage{amsmath,amsfonts,bm}

\def\eqref#1{equation~\ref{#1}}

\def\1{\bm{1}}

\DeclareMathAlphabet{\mathsfit}{\encodingdefault}{\sfdefault}{m}{sl}
\SetMathAlphabet{\mathsfit}{bold}{\encodingdefault}{\sfdefault}{bx}{n}

\usepackage{graphicx}
\usepackage{booktabs}
\usepackage{colortbl}
\usepackage{multirow}
\usepackage{amssymb}
\DeclareMathVersion{framework}
\SetSymbolFont{operators}{framework}{OT1}{ptm}{m}{n}
\SetSymbolFont{letters}{framework}{OML}{cmm}{m}{it}
\SetSymbolFont{symbols}{framework}{OMS}{cmsy}{m}{n}
\SetSymbolFont{largesymbols}{framework}{OMX}{cmex}{m}{n}
\SetMathAlphabet{\mathrm}{framework}{OT1}{ptm}{m}{n}
\SetMathAlphabet{\mathit}{framework}{OT1}{ptm}{m}{it}
\SetMathAlphabet{\mathbf}{framework}{OT1}{ptm}{b}{n}
\usepackage{subcaption}
\usepackage{tikz}
\usetikzlibrary{arrows.meta,positioning,calc,fit,backgrounds,shapes.misc}
\definecolor{GSrow}{HTML}{EDF3FC}
\usepackage{enumitem}
\usepackage{etoolbox}
\newcommand{\GSrowstretch}{1.18}\AtBeginEnvironment{table}{\renewcommand{\arraystretch}{\GSrowstretch}}\AtBeginEnvironment{table*}{\renewcommand{\arraystretch}{\GSrowstretch}}   %

\newcommand{\method}{\textsc{GateSight}}
\newcommand{\gdn}{GDN}

\newcommand{\U}{U}

\title{\method: Recurrent Memory Anticipates Attention Memory\\for Query-Agnostic KV Eviction in Hybrid LLMs}

\author{Junjie Li \qquad Jiong Lou \qquad Jie Li \\
  Shanghai Jiao Tong University}
\hypersetup{pdftitle={GateSight: Recurrent Memory Anticipates Attention Memory for Query-Agnostic KV Eviction in Hybrid LLMs}, pdfauthor={Junjie Li, Jiong Lou, Jie Li}}

\begin{document}
\maketitle

\begin{abstract}
While hybrid language models improve efficiency by replacing most attention layers with fixed-state Gated DeltaNet (\gdn) blocks, their remaining attention key--value (KV) cache still grows linearly, bottlenecking long-context serving. Evicting unimportant KV entries mitigates this bottleneck, yet shared-prefix reuse and long-chain reasoning require estimating importance before future queries or reasoning steps arrive. Designed for full-attention models, existing query-agnostic methods score attention layers in isolation and overlook recurrent \gdn\ signals: training-free heuristics lose semantic salience, while prompt-reconstruction methods incur high overhead and underweight reasoning context. Analyzing leave-one-out continuation utility, we find that recurrent memory anticipates attention needs: native \gdn\ write, decay, and value-norm signals nonlinearly predict future KV utility without positional features. We propose \method, which scores KV entries in linear time solely from preceding \gdn\ signals via lightweight, position-free MLPs trained once on generic text. Across 27B and 125B Qwen3.8 hybrids on 262K prefill and 32K reasoning benchmarks, \method\ retains up to $98.8\%$ of full-cache semantic accuracy at a $10\%$ budget, outperforms reconstruction baselines by $6.7$--$21.8$ points on multi-turn QA and math reasoning (reaching $90.6\%$ on AIME24/25), and boosts 260K decoding throughput by $5.2\times$--$12.2\times$ at under $1.4\%$ scoring latency.
\end{abstract}

\section{Introduction}
\label{sec:intro}

\begin{figure}[t]
\centering
\begingroup%
\mathversion{framework}%
\definecolor{GSblue}{HTML}{1B56A6}%
\definecolor{GSamber}{HTML}{D46A1E}%
\definecolor{GSred}{HTML}{B83B3B}%
\definecolor{GSink}{HTML}{0B0B0B}%
\begin{tikzpicture}[
  x=1pt, y=1.121pt,
  font=\fontsize{7}{8.2}\selectfont, text=GSink,
  >={Stealth[length=3.6pt,width=3.2pt]},
  colhdr/.style={anchor=base, font=\fontsize{6.8}{8}\selectfont, text=black!62, inner sep=0pt},
  rowtitle/.style={anchor=base west, inner sep=0pt, outer sep=0pt, font=\fontsize{7.4}{8.6}\selectfont\bfseries},
  hair/.style={draw=black!30, line width=0.3pt},
  blk/.style={rounded corners=1pt, inner sep=0pt, outer sep=0pt,
    minimum width=38pt, minimum height=11pt, font=\fontsize{7}{8.2}\selectfont},
  gdnidle/.style={blk, draw=black!32, dash pattern=on 1.8pt off 1.2pt, fill=white, text=black!45},
  gdnact/.style={blk, draw=GSblue, fill=GSblue!20, line width=0.6pt,
    font=\fontsize{7}{8.2}\selectfont\bfseries, text=GSblue!85!black},
  attn/.style={blk, draw=GSamber, fill=GSamber!20, line width=0.6pt},
  sm/.style={inner sep=0pt, outer sep=0pt, font=\fontsize{6.6}{7.7}\selectfont},
  lbl/.style={sm, anchor=base},
  sig/.style={->, line width=0.8pt},
  nxt/.style={->, draw=black!55, line width=0.6pt},
  arc/.style={draw=GSamber!85!black, line width=0.45pt},
  evict/.style={draw=black!32, fill=white, line width=0.35pt},
  keep/.style={draw=GSamber!80!black, fill=GSamber, line width=0.35pt},
  need/.style={draw=GSink, line width=0.9pt},
  pics/strip/.style 2 args={code={
    \foreach \k in {0,...,13} {\path[evict] ({140+5.3*\k}, {#1-2.17}) rectangle ++(4.5, 4.34);}
    \foreach \k in {#2} {\path[keep] ({140+5.3*\k}, {#1-2.17}) rectangle ++(4.5, 4.34);}
    \path[need] (166.5, {#1-2.17}) rectangle ++(4.5, 4.34);
  }},
  pics/lost/.style={code={
    \draw[GSred, line width=0.9pt, line cap=round] (167.4, {#1-1.3}) -- (170.1, {#1+1.3}) (167.4, {#1+1.3}) -- (170.1, {#1-1.3});
  }},
  pics/gdn/.style 2 args={code={
    \foreach \d in {2.6, 1.3} {\node[#2, text opacity=0] at ({24+\d}, {#1+\d}) {};}
    \node[#2] at (24, #1) {GDN};
  }},
  pics/unused/.style={code={
    \draw[draw=black!36, line width=0.5pt, dash pattern=on 1.6pt off 1.2pt] (46.6, #1) -- (53.4, #1);
    \draw[draw=black!36, line width=0.5pt] (53.4, {#1-2}) -- (53.4, {#1+2});
    \node[sm, anchor=west, text=black!48] at (56, #1) {\itshape unused};
  }}
]

\fill[GSrow] (0, 0.3) rectangle (219, 50);
\draw[draw=GSink, line width=0.5pt] (0, 151.2) -- (219, 151.2);
\draw[hair] (0, 101.5) -- (219, 101.5);
\draw[hair] (0, 50) -- (219, 50);
\draw[draw=GSink, line width=0.5pt] (0, 0.3) -- (219, 0.3);
\node[colhdr] at (24, 154.7) {Memory read};
\node[colhdr] at (96, 154.7) {Scorer};
\node[colhdr] at (176.7, 154.7) {Kept KV entries};

\node[rowtitle] at (5, 142.4) {(a) Heuristics};
\pic {gdn={130.6}{gdnidle}};
\pic {unused=130.6};
\node[attn] at (24, 114.6) {Attn KV};
\draw[sig, draw=GSamber] (43.6, 114.6) -- (82, 114.6);
\node[lbl] at (62.6, 117.6) {keys / tail};
\node[rounded corners=1pt, inner sep=0pt, draw=GSink, fill=white, line width=0.45pt,
  minimum width=26pt, minimum height=11pt, font=\fontsize{6.8}{7.8}\selectfont] at (96, 114.6) {probe};
\draw[nxt] (109.6, 114.6) -- (137.8, 114.6);
\pic {strip={114.6}{8,10,11,12,13}};
\pic {lost=114.6};
\draw[arc, line width=0.5pt] (203.6, 117.6) -- (203.6, 119.2) -- (213.4, 119.2) -- (213.4, 117.6);
\draw[arc] (208.5, 119.2) .. controls (206.9, 123) and (202.3, 123) .. (200.55, 117.4);
\draw[arc] (208.5, 119.2) .. controls (205.9, 125.8) and (197.9, 125.8) .. (195.25, 117.4);
\draw[arc] (208.5, 119.2) .. controls (203.5, 129.4) and (190, 129.4) .. (184.65, 117.4);
\node[lbl, anchor=base east] at (213.6, 130.2) {tail window};
\draw[draw=GSink, line width=0.4pt] (168.75, 117.3) -- (168.75, 122.8);
\node[lbl] at (166.5, 124.6) {salient};

\node[rowtitle] at (5, 91) {(b) Reconstruction};
\pic {gdn={79.2}{gdnidle}};
\pic {unused=79.2};
\node[attn] at (24, 63.2) {Attn\,/\,$h_\ell$};
\draw[sig, draw=GSamber] (43.6, 63.2) -- (69.8, 63.2);
\node[rounded corners=1pt, inner sep=0pt, draw=GSink, fill=white, line width=0.45pt,
  minimum width=51pt, minimum height=16.4pt] at (96.5, 63.2) {};
\foreach \i in {0,...,4} {\foreach \j in {0,...,4} {
  \path[draw=black!28, fill=white, line width=0.3pt] ({74+2.2*\i}, {57.9+2.12*\j}) rectangle ++(2.2, 2.12);}}
\foreach \i in {0,...,4} {\path[draw=GSamber!80!black, fill=GSamber!85, line width=0.3pt] ({74+2.2*\i}, {57.9+2.12*(4-\i)}) rectangle ++(2.2, 2.12);}
\node[sm, align=center, font=\fontsize{6.4}{7.2}\selectfont] at (103.6, 63.2) {extra pass /\\wide MLP};
\draw[nxt] (122.6, 63.2) -- (137.8, 63.2);
\pic {strip={63.2}{0,1,2,9,10,11}};
\pic {lost=63.2};
\foreach \k in {0,...,13} {\path[draw=black!28, fill=black!9, line width=0.3pt] ({140+5.3*\k}, 70.4) rectangle ++(4.5, 4.34);}
\foreach \k in {0,1,2,9,10,11} {\draw[draw=GSamber!85!black, line width=0.5pt] ({142.25+5.3*\k}, 65.6) -- ({142.25+5.3*\k}, 70.2);}
\node[lbl] at (176.7, 77.8) {prompt self-copy};

\node[rowtitle] at (5, 39.6) {(c) \textsc{GateSight}: cross-memory gating};
\pic {gdn={27.8}{gdnact}};
\node[attn] at (24, 11.8) {Attn KV};
\draw[->, draw=GSamber!75!black, line width=0.45pt, dash pattern=on 2.2pt off 1.4pt] (43.6, 11.8) -- (137.8, 11.8);
\draw[sig, draw=GSblue, line width=0.9pt] (46.6, 27.8) -- (83.6, 27.8);
\node[lbl] at (64.6, 30.8) {$\beta,\, g,\, \|v\|$};
\node[rounded corners=1pt, inner sep=0pt, fill=GSblue, minimum width=24pt,
  minimum height=11pt, font=\fontsize{6.8}{7.8}\selectfont\bfseries, text=white] at (96, 27.8) {MLP};
\node[lbl] at (96, 16.4) {position-free};
\draw[sig, draw=GSblue, line width=0.9pt] (108.6, 27.8) -- (137.8, 27.8);
\pic {strip={11.8}{1,5,8,12}};
\foreach \k/\h in {0/2.6, 2/2.0, 3/3.4, 4/2.4, 6/3.6, 7/2.2, 9/3.0, 10/2.4, 11/3.3, 13/2.1} {
  \fill[GSblue!38] ({140.9+5.3*\k}, 14.9) rectangle ++(2.7, {0.82*\h});
}
\foreach \k/\h in {1/8.6, 5/12.4, 8/9.6, 12/8.0} {
  \fill[GSblue] ({140.9+5.3*\k}, 14.9) rectangle ++(2.7, {0.82*\h});
}
\node[lbl] at (176.7, 29.2) {future utility $\hat{\U}$};

\pgfresetboundingbox
\path[use as bounding box] (0, 0) rectangle (219, 162);
\end{tikzpicture}%
\mathversion{normal}%
\endgroup%
\caption{Query-agnostic KV eviction paradigms in hybrid LLMs (framed cell: the entry a later query needs). \textbf{(a)}~Heuristics read only attention keys or the prompt tail. \textbf{(b)}~Reconstruction supervises prompt self-copying with an extra scoring pass or a wide MLP. \textbf{(c)}~\method\ predicts future utility $\hat{\U}$ from native \gdn\ signals.}
\label{fig:paradigms}
\end{figure}

Extending the context length of large language models (LLMs) is critical for long-document comprehension~\citep{hsieh2024ruler}, shared-context interaction~\citep{li2025scbench}, and extended chain-of-thought reasoning~\citep{cai2025rkv}, yet key--value (KV) caches grow linearly with sequence length and bottleneck serving~\citep{kwon2023vllm}. Recent hybrid LLMs alleviate much of this cost by replacing most attention layers with Gated DeltaNet (\gdn) blocks or variants that compress history into fixed-size matrix states~\citep{yang2025gated,qiu2026qwen38next,kimi2025linear,gu2025jetnemotron,merrill2026olmohybrid}, while retaining a minority of attention layers whose KV cache still scales linearly. In Qwen3.8-27B~\citep{qwen2026qwen38_27b}, all 48 \gdn\ states occupy a fixed 144~MB in FP32, whereas the BF16 KV cache of the 16 full-attention layers consumes 16~GB at 256K tokens ($114\times$ all \gdn\ states) and 64~GB at 1M, capping a 141~GB H200 GPU at four concurrent 260K requests.

KV eviction mitigates this bottleneck by pruning low-importance entries; whereas standard policies rely on observed query attention~\citep{zhang2023h2o,liu2023scissorhands}, shared-prefix reuse~\citep{li2025scbench,kim2025kvzip} and long-chain reasoning~\citep{cai2025rkv} require \emph{query-agnostic} scoring before future queries or reasoning steps arrive. Applying full-attention rules to each hybrid attention layer in isolation creates an architectural mismatch (Figure~\ref{fig:paradigms}a,b): \gdn\ blocks already absorb local context into recurrent states while emitting rich write and decay dynamics across depth, leaving the few attention layers to retrieve exact long-range entities and premises. Blind to this division of labor and preceding \gdn\ signals, training-free heuristics probe only short prompt tails~\citep{li2024snapkv} or static intra-layer Q/K statistics~\citep{devoto2024l2norm,park2025keydiff,mao2026triattention}, losing semantic salience under tight budgets. Meanwhile, prompt-reconstruction methods supervise verbatim self-copying via a costly second pass in KVzip~\citep{kim2025kvzip}---inapplicable to online decoding---or its hidden-state MLP distillation KVzap~\citep{jegou2026kvzap}, incurring high scoring overhead while misallocating scarce KV budget to literal repetition rather than continuation context where fixed-size \gdn\ states are lossy.

We instead ask whether a hybrid model's recurrent memory already anticipates which KV entries its attention memory will later need (Figure~\ref{fig:paradigms}c). Our central hypothesis is that as preceding \gdn\ layers process the shared residual stream, they modulate write gates, decay gates, and write magnitudes when salient entities, premises, or state transitions arrive, flagging the exact tokens full attention must later retrieve once fixed-size states suffer interference. Measuring exact per-entry KV leave-one-out (LOO) output perturbations over query-free continuations on Qwen3.8-27B establishes three findings (\S\ref{sec:obs}): (1)~\gdn\ head half-lives span orders of magnitude and remain anchored by learned parameters across documents (\S\ref{sec:obs1}); (2)~native prefill write, decay, and value-norm signals anticipate future KV utility across domains and lengths, with slow-head writes more than doubling correlation over all-head averaging (\S\ref{sec:obs2}); and (3)~this signal reduces neither to a single oracle head nor to positional bias, but is instead distributed across the preceding \gdn\ layers and nonlinear in the high-utility tail (\S\ref{sec:obs3}).

Guided by these three findings, we propose \method, a query-agnostic KV eviction framework driven by cross-memory \gdn\ signals (\S\ref{sec:method}). \method\ trains lightweight, position-free per-head MLPs once on 1K-token generic excerpts against self-supervised LOO continuation utility, scoring entries in $O(n)$ time solely from preceding \gdn\ signals without materializing attention matrices (\S\ref{sec:estimator}). It pairs these scores with hyperparameter-free bounded gate-proportional layer budgets derived from offline attention output gates (\S\ref{sec:allocation}), enforcing a minimum per-layer floor while keeping per-layer head counts uniform for kernel-free page reclamation in vLLM PagedAttention~\citep{kwon2023vllm} (\S\ref{sec:deploy}).

Evaluated on dense-GQA Qwen3.8-27B and 125B sparse-MoE Qwen3.8-Flash-Next\footnote{\url{https://github.com/hideinsewer/GateSight} (code, vLLM plugins, and trained estimators).} across 262K prefill suites (SCBench, SQuAD, GSM8K) and on 32K reasoning decoding (AIME24/25), \method\ consistently outperforms heuristics, KVzap, and two-pass KVzip: it retains $96.1\%$--$98.8\%$ of FullKV accuracy on SCBench semantic tasks at $p=0.1$, improves QA and math means by $6.7$--$21.8$ points over reconstruction baselines, and reaches $90.6\%$ on AIME24/25 ($+10.0$ over KVzap at a 2K cap), while scoring 256K tokens in $0.200$--$0.254$~s and boosting 260K H200 throughput by $5.2\times$--$12.2\times$. Our contributions are:
\begin{itemize}[leftmargin=1.1em,nosep,topsep=2pt,itemsep=1pt]
\item We show that under parameter-anchored head timescales, native prefill \gdn\ signals nonlinearly anticipate LOO continuation utility without positional bias (\S\ref{sec:obs}).
\item We propose \method, which pairs position-free per-head \gdn\ utility MLPs with bounded gate-proportional layer budgets aligned to PagedAttention block tables (\S\ref{sec:method}).
\item We demonstrate across 27B and 125B hybrids that \method\ outperforms heuristics and two-pass or distilled reconstruction baselines in accuracy, scoring latency, and decoding throughput (\S\ref{sec:exp}).
\end{itemize}

\section{Preliminaries and Utility Formulation}
\label{sec:background}

\paragraph{Hybrid \gdn--Attention Architecture and Native Signals.}
Qwen3.8-27B interleaves 48 \gdn\ layers (3:1) with $L_{\mathrm{att}} = 16$ full-attention layers across 64 layers~\citep{qwen2026qwen38_27b}, while sparse-MoE Qwen3.8-Flash-Next pairs 36 \gdn\ with $L_{\mathrm{att}} = 12$ sparse-attention layers across 48 layers (Appendix~\ref{app:flashnext}). Each attention layer applies a sigmoid output gate before its output projection~\citep{qiu2025gated,qiu2026qwen38next}. Each \gdn\ layer maintains 48 matrix states $S_t \in \mathbb{R}^{128\times128}$ (three value heads per key head), updated per value head via the gated delta rule~\citep{schlag2021linear,yang2024parallelizing,yang2025gated} as follows:
\begin{equation}
S_t = \alpha_t (I - \beta_t k_t k_t^{\top}) S_{t-1} + \beta_t k_t v_t^{\top},
\label{eq:gdn}
\end{equation}
with $\alpha_t = \exp(g_t)$, write gate $\beta_t \in (0,1)$, log-decay $g_t \le 0$, unit-norm key $k_t \in \mathbb{R}^{128}$ ($\|k_t\| = 1$, $\ell_2$ norm), and value $v_t \in \mathbb{R}^{128}$. Because prefill already materializes $\beta_t$, $g_t$, and $v_t$ while $\|k_t\| = 1$ fixes key scale, extracting $(\beta_t, g_t, \|v_t\|)$ requires only a lightweight vector norm with no extra model forward pass.

\paragraph{KV Cache and Two-Axis Eviction.}
Each full-attention layer $\ell$ of Qwen3.8-27B uses grouped-query attention~\citep{ainslie2023gqa} with $H_q=24$ query heads and $H_{\mathrm{kv}} = 4$ KV heads ($d = 256$; $4$~KiB/token per layer, $64$~KiB/token model-wide in BF16), storing $\mathcal{C}_{\ell,h}=\{(k_{\ell,h,i},v_{\ell,h,i})\}_{i=1}^{n}$. Evicting $(k_{\ell,h,i},v_{\ell,h,i})$ removes position $i$ from all six query heads in group $\mathcal{Q}(h)$ while leaving \gdn\ states intact. Given kept fraction $p$ and protected prefix $n_{\mathrm{prot}} \le 256$, prefill eviction selects $\mathcal{I}_{\ell,h} \subseteq \{n_{\mathrm{prot}}+1,\dots,n\}$ under $\sum_{\ell,h}|\mathcal{I}_{\ell,h}| \le B = L_{\mathrm{att}} H_{\mathrm{kv}} \mathrm{round}\big(p(n-n_{\mathrm{prot}})\big)$; online decoding periodically enforces a fixed cap (\S\ref{sec:decode}), decomposing both regimes into two complementary axes: within-layer entry selection (\S\ref{sec:estimator}) and cross-layer budget allocation (\S\ref{sec:allocation}).

\paragraph{Query-Free Continuation Utility.}
To define offline entry importance before user queries arrive (never computed at serving time), we sample $M=4$ self-continuations of up to 128 tokens (steps indexed by $t \in \mathcal{T}$), providing a self-supervised proxy for discourse and reasoning retrieval rather than verbatim copying (\S\ref{sec:exp}). Fixing $(\ell,h)$ (subscripts omitted) and holding continuation queries and remaining entries fixed, deleting entry $i$ shifts the ungated projected output of the attention layer at each continuation step $t \in \mathcal{T}$, which we obtain in closed form:
\begin{align}
\Delta r(t,i) &= \!\!\sum_{j\in\mathcal{Q}(h)}\!\! \frac{a_{j}(t,i)}{1-a_{j}(t,i)} W_O^{(j)}\big(v_i - o_j(t)\big), \nonumber \\
\U(i) &= \max_{t \in \mathcal{T}} \|\Delta r(t,i)\|,
\label{eq:utility}
\end{align}
where $a_j(t,i)$ is the attention weight, $o_j(t)$ the head output, $W_O^{(j)}(v_i - o_j(t))$ discounts high-attention entries near $o_j(t)$, and $\max_{t \in \mathcal{T}}$ retains sparse retrieval peaks. Omitting the output gate preserves within-head rankings ($\rho=0.995$; Appendix~\ref{app:gate}) and enables single-pass evaluation matching explicit deletion within $1.4\times10^{-5}$; as a first-order proxy, $\U(i)$ matches multi-entry Shapley selection~\citep{shapley1953value} under joint eviction (Appendix~\ref{app:label}).

\section{Motivating Analysis: Recurrent Memory Anticipates Attention}
\label{sec:obs}

\begin{figure}[t]
\centering
\includegraphics[width=\columnwidth]{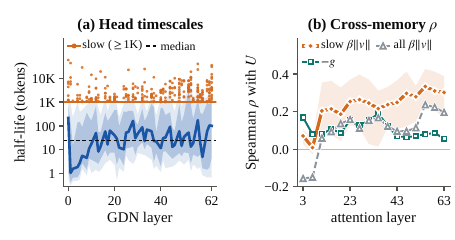}
\caption{\gdn\ timescales and cross-memory correlation on Qwen3.8-27B. \textbf{(a)}~Head half-lives per \gdn\ layer (line: median; bands: 25--75th and 10--90th percentiles; dots: slow heads, $\ge 10^3$~tokens). \textbf{(b)}~Spearman $\rho$ between native signals and future utility $\U$ per attention layer (band: range over the nine domains).}
\label{fig:obs12}
\end{figure}

To test whether recurrent memory anticipates attention needs, we instrument Qwen3.8-27B across 2,428 documents (750--1,250 tokens) from nine Nemotron subsets~\citep{nvidia2025nemotronsample} spanning web text, QA, math, and code (Appendix~\ref{app:obs}), recording prefill \gdn\ signals $(\beta, g, \|v\|)$ alongside the offline utility $\U(i)$ of every KV entry across all 16 attention layers ($\ell \in \{3, 7, \dots, 63\}$; Figure~\ref{fig:obs12}, Table~\ref{tab:obs3}).

\subsection{Observation 1: \gdn\ Head Timescales Are Structured and Parameter-Anchored}
\label{sec:obs1}

\gdn\ heads form a multi-timescale hierarchy anchored by model parameters rather than input text. Measuring each head's retention horizon by its document-mean half-life $\ln 2 / (-\bar{g})$ ($\bar{g} < 0$), every \gdn\ layer spans fast ($<5$ tokens) through slow ($\ge 10^3$ tokens) heads around a 24-token median (Figure~\ref{fig:obs12}a). Across random pairs of 240 documents, half-life rankings over all 2,304 value heads agree at Spearman $\rho = 0.990$--$0.997$ (10th--90th percentiles), and the zero-input decay parameters of the checkpoint alone predict empirical half-lives at $\rho = 0.942$ (Figure~\ref{fig:obs1}; Appendix~\ref{app:obs}). Single-step write magnitude $\beta\|v\|$ ($= \|\beta k v^{\top}\|_F$ under $\|k\| = 1$) is similarly structured: head identity explains $71.7\%$ of the variance in $\log(\beta\|v\|)$ versus $10.2\%$ for token identity, and slow heads write far more selectively than fast heads (within-layer $\rho = -0.66$ between half-life and mean $\beta\|v\|$), allowing a learned estimator to assign fixed weights to individual heads, independent of the document being processed.

\subsection{Observation 2: Native \gdn\ Write and Decay Signals Anticipate Future KV Utility}
\label{sec:obs2}

\begin{table}[t]
\centering
\footnotesize
\setlength{\tabcolsep}{2.0pt}
\begin{tabular*}{\columnwidth}{@{\extracolsep{\fill}}lccc}
\toprule
Predictor & $\rho$ ($\uparrow$) & Ov@10 & Cov@10 \\
\midrule
\multicolumn{4}{@{}l}{\textit{Untrained signals}} \\
Random & $+0.000$ & 10.0 & 19.0 \\
Age (older first) & $+0.071$ & 19.2 & 31.2 \\
Slow-head $\beta\|v\|$ & $+0.229$ & 18.2 & 30.6 \\
Single-head oracle$^*$ & $+0.232$ & 22.6 & 36.0 \\
\midrule
\multicolumn{4}{@{}l}{\textit{Learned predictors (LOCO)}} \\
Age MLP (3 feat.) & $+0.149$ & 17.9 & 29.6 \\
Ridge (864 \gdn) & {\boldmath$+0.527$} & 36.4 & 52.6 \\
\rowcolor{GSrow}\method\ MLP (864 \gdn) & $+0.523$ & \textbf{38.1} & \textbf{54.6} \\
\bottomrule
\end{tabular*}

\caption{Predicting utility $\U$ on Qwen3.8-27B (2,428 documents, 16 attention layers). $\rho$: Spearman; Ov@10/\allowbreak Cov@10: top-10\% index overlap and utility-mass coverage (\%); $^*$: selected in-sample. In all tables, \textbf{bold} marks the best result and shading marks ours.}
\label{tab:obs3}
\end{table}

Averaged over up to six preceding \gdn\ layers (\S\ref{sec:estimator}), every native signal ($\beta$, $-g$, $\|v\|$, $\beta\|v\|$) correlates positively with $\U$ across all nine domains (Appendix~\ref{app:obs}). Restricting window-averaged $\beta\|v\|$ to slow heads (from the 293 global heads with median half-life $\ge 10^3$ tokens) more than doubles mean correlation over all-head $\beta\|v\|$ ($+0.10\to+0.23$), sustaining $\rho = +0.19$--$+0.33$ from $\ell = 11$ onward once recurrent context accumulates. Decay magnitude $-g$ likewise correlates positively across all 16 layers ($\rho = +0.06$--$+0.19$; Figure~\ref{fig:obs12}b), showing that high-utility KV entries arrive where \gdn\ heads clear prior state and write salient updates. This fixed slow-head set transfers zero-shot to 2K--8K documents and RULER-8K~\citep{hsieh2024ruler} ($\rho = +0.30$--$+0.34$; Table~\ref{tab:transfer}).

\begin{figure*}[t]
\centering
\begingroup%
\mathversion{framework}%
\definecolor{GSblue}{HTML}{1B56A6}%
\definecolor{GSamber}{HTML}{D46A1E}%
\definecolor{GSink}{HTML}{0B0B0B}%
\colorlet{GSpanel}{black!5}%
\begin{tikzpicture}[
  x=1pt, y=1.272pt,
  font=\fontsize{7.6}{9}\selectfont, text=GSink,
  >={Stealth[length=3.4pt,width=2.9pt]},
  note/.style={align=center, inner sep=0pt, outer sep=0pt},
  small/.style={font=\fontsize{6.8}{8}\selectfont},
  heading/.style={note, anchor=base west, font=\fontsize{8.2}{10}\selectfont\bfseries},
  panel/.style={rounded corners=2pt, fill=GSpanel},
  flow/.style={->, draw=black!62, line width=.6pt, shorten <=1.5pt, shorten >=1.5pt},
  link/.style={->, line width=1.3pt, >={Stealth[length=4.4pt,width=4.4pt]}},
  layer/.style={rounded corners=1pt, inner sep=0pt, outer sep=0pt, minimum width=35pt, minimum height=8.6pt,
    line width=.5pt, font=\fontsize{6.6}{7.6}\selectfont},
  gdnl/.style={layer, draw=GSblue, fill=GSblue!18},
  attl/.style={layer, draw=GSamber, fill=GSamber!22},
  cell/.style={inner sep=0pt, outer sep=0pt, minimum width=5.4pt, minimum height=5.4pt},
  kept/.style={fill=GSamber, draw=GSamber!80!black, line width=.35pt},
  evct/.style={fill=white, draw=black!38, line width=.35pt},
  cut/.style={draw=black!60, line width=.45pt, dash pattern=on 1.6pt off 1.1pt}
]

\foreach \xa/\xb/\title in {0.3/114/(1) Prefill once, 122/240/(2) Estimate utility,
    248/352/(3) Allocate \& select, 360/454.9/(4) Compact \& serve} {
  \path[panel] (\xa, 30) rectangle (\xb, 160.6);
  \node[heading] at ({\xa+5.4}, 149.9) {\title};
}

\node[note, small] at (66.1, 140.6) {$\beta,\ g,\ \|v\|$ per token, head};
\foreach \j/\lab in {0/1, 3/$i$, 6/$n$} {
  \node[note, small, anchor=base] at ({46.6+6.5*\j}, 128.6) {\lab};
}
\foreach \k/\lab in {0/$\ell{-}7$, 1/$\ell{-}6$, 2/$\ell{-}5$, 4/$\ell{-}3$, 5/$\ell{-}2$, 6/$\ell{-}1$} {
  \node[gdnl] at (22.5, {121-11.4*\k}) {GDN \lab};
  \foreach \j in {0,1,2,4,5,6} {
    \node[cell, fill=GSblue!26] at ({46.6+6.5*\j}, {121-11.4*\k}) {};
  }
  \node[cell, fill=GSblue] at (66.1, {121-11.4*\k}) {};
}
\node[attl, draw=GSamber!45, fill=GSamber!8, text=black!55] at (22.5, 86.8) {Attn $\ell{-}4$};
\foreach \j in {0,...,6} {
  \node[cell, fill=GSamber!13] at ({46.6+6.5*\j}, 86.8) {};
}
\node[attl] (L7) at (22.5, 41.2) {Attn $\ell$};
\foreach \j in {0,1,2,4,5,6} {
  \node[cell, fill=GSamber!32] at ({46.6+6.5*\j}, 41.2) {};
}
\node[cell, fill=GSamber] at (66.1, 41.2) {};
\draw[draw=GSblue, line width=.6pt, rounded corners=1.5pt] (89.8, 125.4) -- (93.2, 125.4) -- (93.2, 48.2) -- (89.8, 48.2);
\draw[link, draw=GSblue] (93.2, 86.8) -- (128, 86.8);
\node[note, small, anchor=base] at (103.4, 90.8) {$W(\ell)$};

\foreach \s in {0,...,5} {
  \fill[GSblue] (129.4, {58.5+9.6*\s}) rectangle ++(10, 8.6);
}
\node[note, anchor=base] at (134.4, 118.8) {$x_i^{(\ell)}$};
\path[fill=GSblue, rounded corners=1.2pt] (149.4, 115.1) -- (181, 99.4) -- (181, 74.2) -- (149.4, 58.5) -- cycle;
\node[note, font=\fontsize{7.4}{8.6}\selectfont\bfseries, text=white] at (164.6, 86.8) {MLP};
\draw[flow, draw=GSblue, shorten <=.8pt, shorten >=.8pt] (139.4, 86.8) -- (149.4, 86.8);
\node[circle, inner sep=0pt, minimum size=5pt, fill=GSblue] (out) at (198.8, 86.8) {};
\draw[flow, draw=GSblue, shorten >=.5pt] (181, 86.8) -- (out.west);
\node[note, small, anchor=base] at (188.6, 90.3) {$\sigma$};
\node[note, anchor=west] at (203.6, 86.8) {$\hat{\U}_i^{(\ell,h)}$};
\node[note, small, anchor=base] at (181, 41.6) {position-free,\enspace $O(n)$,\enspace per head};
\draw[draw=black!45, line width=.4pt] (191.4, 119) -- (238, 119);
\foreach \k/\v in {0/4.0,2/2.7,3/11.7,4/5.0,5/3.1,6/4.3,7/12.8,8/3.5,9/10.3,10/11.3} {
  \fill[GSblue!38] ({193.5+4*\k}, 119) rectangle ++(2.6, \v);
}
\fill[GSblue] (197.5, 119) rectangle ++(2.6, 11);
\draw[flow, draw=GSblue, shorten <=0pt, shorten >=.3pt] (198.8, 89.6) -- (198.8, 119);
\node[note, small, anchor=base west] at (203.6, 107.6) {entries $1..n$};

\draw[link, draw=GSblue] (238.8, 124.4) -- (254.4, 124.4);
\foreach \k/\v in {1/11,3/11.7,7/12.8,9/10.3,10/11.3} {
  \fill[GSblue] ({256.2+6.4*\k}, 119.2) rectangle ++(3.6, \v);
  \path[kept] ({255.7+6.4*\k}, 114.1) rectangle ++(4.6, 4.25);
}
\foreach \k/\v in {0/4.0,2/2.7,4/5.0,5/3.1,6/4.3,8/3.5} {
  \fill[GSblue!38] ({256.2+6.4*\k}, 119.2) rectangle ++(3.6, \v);
  \path[evct] ({255.7+6.4*\k}, 114.1) rectangle ++(4.6, 4.25);
}
\draw[cut] (254, 127.1) -- (326, 127.1);
\node[note, small, anchor=base] at (290.3, 138.6) {top-$K_\ell$ kept, rest evicted};
\node[note, draw=GSamber, fill=white, rounded corners=1pt, inner xsep=4pt, inner ysep=2.6pt, line width=.5pt] (K)
  at (290.3, 96.2) {$K_\ell \propto w_\ell \;(\geq k_{\min})$};
\draw[->, draw=GSamber, line width=1pt, >={Stealth[length=3.8pt,width=3.6pt]}] (290.3, 102.6) -- (290.3, 113.4);
\draw[->, draw=GSamber, line width=1pt, >={Stealth[length=3.8pt,width=3.6pt]}] (290.3, 68) -- (290.3, 89.6);
\draw[draw=black!45, line width=.4pt] (251.6, 42) -- (329, 42);
\foreach \k/\h in {0/0.75,1/1.06,2/2.06,3/2.29,4/4.0,5/5.82,6/8.51,7/8.45,
    8/8.43,9/8.45,10/9.87,11/6.85,12/8.9,13/12.14,14/16,15/11.68} {
  \fill[GSamber] ({254.3+4.6*\k}, 42) rectangle ++(3.4, {2.4*\h});
}
\draw[cut] (251.6, 59.9) -- (329, 59.9);
\node[note, small, anchor=west] at (330.6, 59.9) {flat};
\node[note, small, anchor=base west] at (253, 72) {$w_\ell$};
\node[note, small, anchor=base west] at (251.6, 33.6) {shallow};
\node[note, small, anchor=base east] at (329, 33.6) {deep};

\draw[link, draw=GSamber] (328.4, 119) -- (366.6, 119);
\draw[draw=GSamber, line width=.6pt, rounded corners=1.5pt] (370.6, 136.4) -- (368, 136.4) -- (368, 101.6) -- (370.6, 101.6);
\node[note, small, anchor=base] at (415, 138.6) {KV pages};
\foreach \y/\lab/\keptlist in {132.5/$\ell_1$/{3}, 123.5/$\ell_2$/{1,4}, 114.5/$\ell_3$/{1,3,4}, 105.5/$\ell_4$/{0,1,3,5}} {
  \node[note, small, anchor=east] at (380, \y) {\lab};
  \fill[GSamber!28] (382, \y-3.2) rectangle (448, \y+3.2);
  \foreach \p in \keptlist {
    \fill[GSamber] ({382+11*\p}, \y-3.2) rectangle ({393+11*\p}, \y+3.2);
  }
  \foreach \p in {1,...,5} {
    \draw[GSpanel, line width=.7pt] ({382+11*\p}, \y-3.2) -- ({382+11*\p}, \y+3.2);
  }
}
\draw[->, draw=black!62, line width=.9pt, >={Stealth[length=3.6pt,width=3.4pt]}] (388, 100.4) -- (388, 89.4);
\node[note, small, anchor=base west] at (393, 92.4) {pack to front};
\foreach \y/\lab/\nk in {84/$\ell_1$/1, 75/$\ell_2$/2, 66/$\ell_3$/3, 57/$\ell_4$/4} {
  \node[note, small, anchor=east] at (380, \y) {\lab};
  \foreach \p in {0,...,5} {                                                  %
    \path[fill=white, draw=black!42, line width=.4pt, dash pattern=on 1.4pt off 1pt]
      ({382.5+11*\p}, \y-2.95) rectangle ({392.5+11*\p}, \y+2.95);
  }
  \fill[GSamber] (382, \y-3.2) rectangle ({382+11*\nk}, \y+3.2);              %
  \foreach \p in {1,...,\nk} {
    \draw[GSpanel, line width=.7pt] ({382+11*\p}, \y-3.2) -- ({382+11*\p}, \y+3.2);
  }
  \path[fill=white, draw=GSamber, line width=.5pt] ({382.6+11*\nk}, \y-2.95) rectangle ({392.4+11*\nk}, \y+2.95);   %
  \node[note, font=\fontsize{6.6}{7.6}\selectfont] at ({387.5+11*\nk}, \y) {$q$};
}
\draw[->, draw=black!62, line width=.9pt, >={Stealth[length=3.6pt,width=3.4pt]}] (437.5, 52.4) -- (437.5, 41.6);
\node[note, small, anchor=base east] at (448, 33.6) {free to block pool};

\draw[->, draw=GSamber, line width=.7pt, rounded corners=3pt, dash pattern=on 2.5pt off 1.5pt]
  (22.5, 37) -- (22.5, 23.6) -- (290.3, 23.6) -- (290.3, 41.6);
\node[note, small, fill=white, inner xsep=3pt, inner ysep=1pt] at (157, 23.6)
  {offline attention output-gate mean $w_\ell$};

\pgfresetboundingbox
\path[use as bounding box] (0, 17) rectangle (455.2, 161);
\end{tikzpicture}%
\mathversion{normal}%
\endgroup%
\caption{\method\ overview. \textbf{(1)}~Prefill emits native \gdn\ signals $(\beta, g, \|v\|)$. \textbf{(2)}~Position-free MLPs estimate per-head utility $\hat{\U}_i^{(\ell,h)}$. \textbf{(3)}~Offline output-gate means $w_\ell$ set bounded layer budgets $K_\ell$. \textbf{(4)}~Kept entries pack into leading vLLM pages ($q$: later tokens); freed pages return to the block pool.}
\label{fig:framework}
\end{figure*}

\subsection{Observation 3: Cross-Memory Signal Is Distributed, Distinct, and Tail-Nonlinear}
\label{sec:obs3}

Individual signals and single heads remain too weak for tight budgets: across the up to $6\times 48 = 288$ (\gdn\ layer, head) cells in the preceding window ($144$ at $\ell = 3$), $40\%$--$70\%$ exceed $|\rho|>0.1$ for $\beta\|v\|$ yet none exceeds $0.39$, and the top layers shift with depth (Figures~\ref{fig:obs2layers}--\ref{fig:obs3cells}). Untrained signals thus recover at most $19.2\%$ of true top-10\% entries (Ov@10), and even an in-sample single-head oracle reaches only $22.6\%$ (Table~\ref{tab:obs3}; Table~\ref{tab:obs3full}).

Concatenating $(\beta, g, \|v\|)$ across this window (\S\ref{sec:estimator}) under 9-subset leave-one-corpus-out (LOCO) evaluation establishes two further properties (Table~\ref{tab:obs3}; Appendix~\ref{app:obs}). First, predictive power comes from cross-memory \gdn\ dynamics rather than positional shortcuts or generic MLP capacity: the position-free \gdn\ MLP reaches $38.1\%$ Ov@10 and captures $54.6\%$ of oracle top-10\% utility mass (Cov@10; $64.0\%$ Cov@20 under heavy-tailed $\U$), versus $17.9\%$ and $29.6\%$ for a matched MLP on three position/age features alone ($+20.2$ Ov@10), and beats matched position-free MLPs on TriAttention ($28.8\%$ Ov@10, $\rho=0.389$) or KeyDiff ($18.1\%$, $\rho=0.213$) in 240-document LOCO ($38.2\%$ Ov@10, $\rho=0.516$; Table~\ref{tab:signal_mlp}). Second, despite similar global $\rho$ to ridge regression ($+0.523$ vs.\ $+0.527$), the nonlinear MLP outperforms ridge in every attention layer ($38.1\%$ vs.\ $36.4\%$ Ov@10; $54.6\%$ vs.\ $52.6\%$ Cov@10; Figure~\ref{fig:obs3layers}), better resolving the high-utility tail governing tight budgets.

\section{\method: Native Gate-Guided KV Eviction}
\label{sec:method}

\method\ (Figure~\ref{fig:framework}) co-designs query-agnostic eviction around native signal complementarity and serving-engine alignment. On the modeling side, \gdn\ signals vary richly across tokens to rank within-layer utility (\S\ref{sec:estimator}), whereas attention output gates span over $20\times$ across layers to set cross-layer budgets (\S\ref{sec:allocation}); on the systems side, pairing head-specific rankings with head-uniform layer counts $K_\ell$ packs retained caches into contiguous PagedAttention pages without head padding (\S\ref{sec:deploy}).

\subsection{Within-Layer Utility Estimator}
\label{sec:estimator}

For attention layer $\ell$ and KV head $h$, \method\ scores entry $i \in \{1,\dots,n\}$ from all $H_g=48$ \gdn\ value heads across a local window $W(\ell)$ of up to six preceding \gdn\ layers, with zero positional features:
\begin{equation}
x_i^{(\ell)} \!=\! \big[\{\beta^{(j)}_{i,m}, g^{(j)}_{i,m}, \|v^{(j)}_{i,m}\|\}_{j\in W(\ell),\,m\le H_g}\big].
\label{eq:features}
\end{equation}
On both backbones, $d_\ell = 432$ at the first attention layer ($|W(\ell)| = 3$) and $d_\ell = 864$ thereafter ($|W(\ell)| = 6$). After training-set standardization, an independent $d_\ell \to 64 \to 1$ GELU--sigmoid MLP $f_\theta^{(\ell,h)}$ per $(\ell,h)$ predicts $\hat{\U}_i^{(\ell,h)} = f_\theta^{(\ell,h)}(x_i^{(\ell)}) \in (0,1)$ in $O(n)$ time without attention matrices, totaling 3.4M parameters across 64 MLPs on Qwen3.8-27B and 1.3M across 24 MLPs on Flash-Next. Even when the best single \gdn\ layer precedes $W(\ell)$, the best in-window layer trails by $<0.07$ in $\rho$ (Figure~\ref{fig:obs2layers}), and sweeping $3$--$12$ layers confirms $|W(\ell)| = 6$ as the accuracy--size knee (Table~\ref{tab:window}; Appendix~\ref{app:ablation}).

\paragraph{Self-Supervised Training and Length Scaling.}
Each $f_\theta^{(\ell,h)}$ is trained with binary cross-entropy against $\U_i^{(\ell,h)}$ (Eq.~\ref{eq:utility}, max-normalized over $i$ per document and KV head) at 500 random positions per document across the 2,428 Nemotron 1K documents (AdamW, $\text{lr}=2\times10^{-3}$, 12 epochs; Appendix~\ref{app:details}), using no queries or task labels. Reading purely local recurrent updates under parameter-anchored timescales (\S\ref{sec:obs1}), the unchanged estimator transfers to unseen 2K--8K documents ($\rho = 0.56$--$0.58$) and RULER-8K ($\rho = 0.39$; Table~\ref{tab:transfer}), as well as 262K prefill and 32K decoding (\S\ref{sec:exp}).

\subsection{Cross-Layer Allocation from Output Gates}
\label{sec:allocation}

Summing all $H_q=24$ query-head outputs into the residual stream scales each layer's effective throughput by the offline query-head gate mean $w_\ell = \frac{1}{H_q}\sum_{j=1}^{H_q}\mathbb{E}[\sigma(z_{\ell,j})]$ over calibration prefill logits $z_{\ell,j}$. Across layers, $w_\ell$ spans $0.022$--$0.459$ ($21.2\times$) on Qwen3.8-27B and $0.177$--$0.418$ ($2.4\times$) on Flash-Next (Appendix~\ref{app:gate}); zeroing shallow layers 3, 7, and 11 leaves RULER-8K at $96.2\%$, whereas also zeroing layer 23 collapses it to $39.1\%$. Given kept fraction $p$ and allocatable budget $B$ (\S\ref{sec:background}), \method\ sets per-layer per-head counts $K_\ell$ via bounded gate-proportional allocation:
\begin{equation}
K_\ell - n_{\mathrm{prot}} = \mathrm{clip}(\lambda w_\ell,\; k_{\min},\; n - n_{\mathrm{prot}}),
\label{eq:alloc}
\end{equation}
solving for $\lambda$ (largest-remainder rounding; Appendix~\ref{app:details}) so $H_{\mathrm{kv}}\sum_\ell (K_\ell - n_{\mathrm{prot}}) = B$ with floor $k_{\min}=4$ and no temperature hyperparameter, guaranteeing a per-layer floor absent from global top-$B$ selection in KVzip~\citep{kim2025kvzip} and KVzap~\citep{jegou2026kvzap} while preserving the shared per-layer block tables of the serving engine (Appendix~\ref{app:gate}).

\begin{figure*}[t]
\centering
\includegraphics[width=\textwidth]{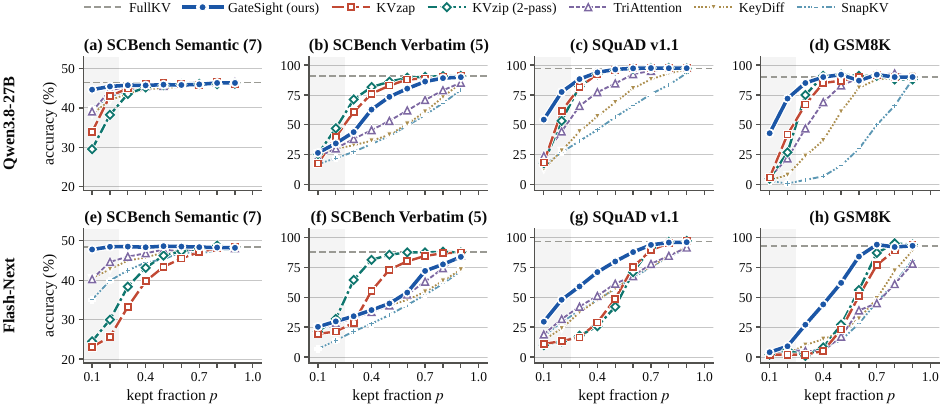}
\caption{Prefill eviction accuracy (\%) vs.\ kept fraction $p$ on Qwen3.8-27B (top) and Qwen3.8-Flash-Next (bottom). Dashed: FullKV; shaded: $p\le 0.2$.}
\label{fig:prefill}
\end{figure*}

\subsection{Inference-Engine Compaction: Dense, Sparse, and Decoding}
\label{sec:deploy}

\paragraph{Dense-GQA Prefill.}
Implemented as a drop-in vLLM 0.28~\citep{kwon2023vllm} plugin with no custom CUDA kernels, \method\ retains $n_{\mathrm{prot}}$ plus the top-$(K_\ell - n_{\mathrm{prot}})$ post-RoPE entries per KV head at prefill end, packs all $K_\ell$ entries in sequence order into leading pages, and returns freed tail blocks to the pool while preserving \gdn\ states for multi-turn reuse.

\paragraph{Sparse-QSA Prefill.}
On Flash-Next, QSA~\citep{qiu2026qwen38next} indexes native 4-token blocks shared by both KV heads; \method\ pools scores via a two-head max and 4-token block average, rounds $K_\ell$ to multiples of four, and copies retained KV blocks alongside compressed-key summaries, reclaiming $(1-p)$ of pages and shrinking QSA indexer scans from $n/4$ to $pn/4$ blocks (details in Appendix~\ref{app:flashnext}).

\paragraph{Online Decoding.}
Because $\hat{\U}_t^{(\ell,h)}$ is static once token $t$ is emitted, decoding buffers only the latest 128-step $(\beta_t, g_t, \|v_t\|)$ chunk for batched scoring and keeps one scalar score per retained entry at periodic compaction, while unpruned \gdn\ states preserve full recurrent history (\S\ref{sec:decode}).

\section{Experiments}
\label{sec:exp}

\subsection{Experimental Setup}
\label{sec:setup}

\paragraph{Models and Benchmarks.}
Using the Nemotron-trained estimators of \S\ref{sec:estimator}, we evaluate dense-GQA Qwen3.8-27B (\S\ref{sec:background}) and 125B sparse-MoE Qwen3.8-Flash-Next~\citep{qiu2026qwen38next} (6B active, FP8, 12 QSA layers; Appendix~\ref{app:flashnext}). For prefill eviction ($p\in\{0.9,\dots,0.1\}$), we evaluate 12-task SCBench~\citep{li2025scbench} up to 262K tokens (official metrics with two uniform parser fixes; Appendix~\ref{app:protocol}), split into 7 semantic comprehension and 5 verbatim lookup tasks, alongside multi-turn SQuAD v1.1~\citep{rajpurkar2016squad} (100 passages, 800 QAs) and zero-shot CoT GSM8K~\citep{cobbe2021gsm8k} (100 setup prefixes $\ge 72$ tokens compressed prior to the question) adapted from KVzip~\citep{kim2025kvzip}, plus RULER-4096~\citep{hsieh2024ruler} in Appendix~\ref{app:ruler}. For online decoding eviction, we evaluate all 60 AIME24/25 problems on Qwen3.8-27B (Flash-Next QSA natively caps attention at 2,048 tokens), compacting 32K \texttt{<think>} trajectories ($T=0.6$, top-$p=0.95$, 2K answer reserve) every 128 steps to a mean per-head cap of $4{,}096$, $3{,}072$, or $2{,}048$ cached entries~\citep{cai2025rkv,mao2026triattention}.

\paragraph{Baselines.}
We compare against training-free heuristics---SnapKV~\citep{li2024snapkv} (32-token tail), TriAttention~\citep{mao2026triattention}, KeyDiff~\citep{park2025keydiff}, and StreamingLLM~\citep{xiao2024streamingllm} (decoding)---and reconstruction baselines: two-pass KVzip~\citep{kim2025kvzip} and single-pass KVzap~\citep{jegou2026kvzap}, trained on the same 2,428 Nemotron documents against $\log\text{KVzip+}$ (52.5M params, validation $R^2=0.784$ on 27B; 15.7M on Flash-Next). All prefill methods run in our shared vLLM \texttt{fork\_session} engine with verified physical page release under paper-faithful layer budgets (\texttt{flat} for heuristics, global top-$B$ for KVzip and KVzap, and bounded gate-proportional allocation for \method; Appendices~\ref{app:protocol} and~\ref{app:details}).

\subsection{Prefill: Continuation vs.\ Reconstruction}
\label{sec:prefill}

Figure~\ref{fig:prefill} plots nine-budget prefill trajectories across both hybrids (Appendix~\ref{app:pertask}), comparing \method\ against heuristic and prompt-reconstruction baselines.

\paragraph{Comprehension, QA, and Reasoning.}
Across all comprehension, QA, and reasoning suites, \method\ outperforms every heuristic, KVzap, and two-pass KVzip, with widest margins at $p\le 0.3$ and under sparse QSA. On 262K SCBench Semantic (7 tasks; Figure~\ref{fig:prefill}a,e), \method\ tracks FullKV down to $p=0.1$, retaining $96.1\%$ on 27B ($44.6\%$ vs.\ $46.4\%$) and $98.8\%$ on Flash-Next ($47.8\%$ vs.\ $48.4\%$), whereas KVzap and KVzip drop to $33.9\%$/$29.5\%$ on 27B and $23.1\%$/$24.5\%$ on Flash-Next (up to $+55.6$ points on \texttt{summary\_with\_needles}). On SQuAD v1.1 (Figure~\ref{fig:prefill}c,g), \method\ attains an $88.9\%$ mean on 27B versus $82.2\%$ (KVzap) and $81.2\%$ (KVzip; $54.4\%$ vs.\ $18.5\%$ and $18.0\%$ at $p=0.1$), and $73.4\%$ on Flash-Next versus $52.8\%$ and $51.6\%$ ($+20.6$ and $+21.8$ points), where reconstruction degrades sharply by $p=0.5$. On GSM8K (Figure~\ref{fig:prefill}d,h), \method\ reaches $82.3\%$ on 27B versus $72.1\%$ and $72.0\%$ ($43.0\%$ vs.\ $6.0\%$ and $5.0\%$ at $p=0.1$), and $56.6\%$ on Flash-Next versus $38.3\%$ and $41.8\%$ ($+18.2$ and $+14.8$ points; $+39.0$ and $+35.0$ at $p=0.5$, before $p\le 0.2$ leaves only $2$--$5$ four-token blocks on $\sim\!100$-token setups).

\begin{figure}[t]
\centering
\includegraphics[width=\columnwidth]{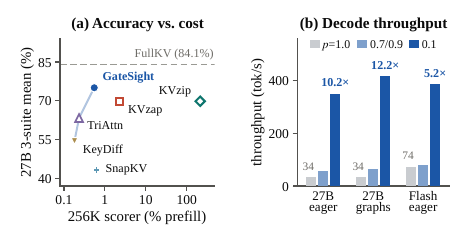}
\caption{Scoring latency and decoding throughput. \textbf{(a)}~Mean accuracy over SCBench, SQuAD, and GSM8K vs.\ 256K scorer latency on Qwen3.8-27B (line: Pareto frontier). \textbf{(b)}~260K H200 decoding throughput at $p=1.0$, $0.7$ (27B) or $0.9$ (Flash-Next), and $0.1$.}
\label{fig:eff}
\end{figure}

\paragraph{Verbatim Lookup and Tight-Budget Collapse.}
On the five verbatim tasks of SCBench (Figure~\ref{fig:prefill}b,f), KVzip and KVzap lead at moderate budgets by supervising literal self-copying ($86.1\%$ and $82.8\%$ vs.\ $73.4\%$ at $p=0.5$ on 27B), lifting 27B 12-task means to $56.2\%$ and $55.2\%$ versus $53.9\%$ for \method, whereas on Flash-Next \method\ surpasses KVzap overall ($49.5\%$ vs.\ $47.8\%$); retraining matched \gdn\ MLPs on reconstruction attention instead of continuation utility recovers part of this verbatim gap on mixed code--key retrieval but not on random-UUID lookup (Table~\ref{tab:supervision}, Appendix~\ref{app:ablation}). At $p=0.1$, however, both native reconstruction baselines drop sharply even on verbatim lookup: on 27B and Flash-Next, \method\ matches or beats KVzap on 11/12 SCBench tasks ($+9.9$ and $+16.9$ overall; $+8.9$ and $+5.8$ verbatim) and KVzip on 10/12 ($+9.2$ and $+15.3$ overall). RULER-4096 follows the same pattern (Appendix~\ref{app:ruler}), with \method\ ranking first on 27B ($92.3\%$ vs.\ $88.6\%$ KVzap and $83.4\%$ KVzip; $+37.7$ and $+47.1$ at $p=0.1$) and surpassing both reconstruction baselines on Flash-Next ($79.4\%$ vs.\ $62.4\%$ and $74.0\%$, and within $0.5$ points of KeyDiff).

\begin{table}[t]
\centering
\footnotesize
\setlength{\tabcolsep}{4.5pt}
\begin{tabular*}{\columnwidth}{@{\extracolsep{\fill}}lccc}
\toprule
Benchmark & Flat & Gate-Prop. & $\Delta$ \\
\midrule
27B SCBench-3 & 53.7 & \textbf{60.6} & $+6.9$ \\
27B SQuAD v1.1 & 85.9 & \textbf{88.9} & $+2.9$ \\
27B GSM8K & 78.2 & \textbf{82.3} & $+4.1$ \\
27B AIME24/25 & 84.4 & \textbf{90.6} & $+6.1$ \\
\midrule
Flash-Next SQuAD & 69.5 & \textbf{73.4} & $+3.8$ \\
Flash-Next GSM8K & 49.9 & \textbf{56.6} & $+6.7$ \\
\bottomrule
\end{tabular*}

\caption{Budget-mean accuracy (\%) under \texttt{flat} vs.\ gate-proportional allocation with fixed \method\ scores ($\Delta$: gain before rounding; details in Table~\ref{tab:ablation}).}
\label{tab:gate_main}
\end{table}

\subsection{Efficiency, Allocation, and 260K Serving}
\label{sec:eff}

\paragraph{Scoring Efficiency.}
Across the three 27B prefill suites (Figure~\ref{fig:eff}a; Appendices~\ref{app:ablation}--\ref{app:cost}), \method\ leads the Pareto frontier at a $75.0\%$ mean ($+5.2$ over KVzap and KVzip at $69.8\%$). In post-prefill scorer latency (unfused hook overhead in Appendix~\ref{app:cost}), \method\ scores 256K tokens in $0.254$~s on 27B ($0.55\%$ of $46.3$~s prefill)---$4.2\times$ faster than KVzap ($1.073$~s, $15.3\times$ larger) and $390\times$ faster than KVzip ($99.2$~s)---and $0.200$~s on Flash-Next ($1.4\%$ of $13.9$~s prefill, vs.\ $0.247$~s and $52.8$~s).

\paragraph{Layer Allocation.}
With \method\ scores fixed (Tables~\ref{tab:gate_main},~\ref{tab:ablation}), uniform \texttt{flat} budgets beat all \texttt{flat} heuristics plus native KVzap/KVzip on SQuAD and GSM8K ($46.2\%$ vs.\ $24.0\%$ TriAttention and $18.5\%$ KVzap on 27B SQuAD at $p=0.1$) and 2K AIME ($81.7\%$ vs.\ $71.7\%$ and $76.7\%$), so gains do not hinge on gate budgets, while gate allocation adds $+2.9$--$6.9$ points across all six suites (beating PyramidKV).

\paragraph{260K Serving.}
Physical vLLM page reclamation at 260K context (Figure~\ref{fig:eff}b; Table~\ref{tab:serving}) raises H200 decoding throughput on 27B by $1.7\times$--$1.9\times$ at $p=0.7$ and by $10.2\times$ ($350.4$~tok/s, eager) to $12.2\times$ ($418.5$~tok/s, CUDA graphs) at $p=0.1$ ($B_{\max}: 4\to 41$), and by $5.2\times$ on Flash-Next ($B_{\max}: 9\to 86$, $388.2$~tok/s), avoiding the $+10.6\%$--$29.5\%$ TPOT penalty of ragged per-head paging (Appendix~\ref{app:details}).

\begin{table*}[t]
\centering
\footnotesize
\setlength{\tabcolsep}{4pt}
\begin{tabular*}{\textwidth}{@{\extracolsep{\fill}}lcccccccccc}
\toprule
 & \multicolumn{3}{c}{AIME24 (30)} & \multicolumn{3}{c}{AIME25 (30)} & \multicolumn{4}{c}{60-Problem Mean} \\
\cmidrule(lr){2-4}\cmidrule(lr){5-7}\cmidrule(lr){8-11}
Method & 4,096 & 3,072 & 2,048 & 4,096 & 3,072 & 2,048 & 4,096 & 3,072 & 2,048 & Mean \\
\midrule
FullKV & \multicolumn{3}{c}{96.7} & \multicolumn{3}{c}{93.3} & \multicolumn{4}{c}{95.0} \\
\midrule
StreamingLLM & 66.7 & 60.0 & 53.3 & 53.3 & 50.0 & 43.3 & 60.0 & 55.0 & 48.3 & 54.4 \\
SnapKV & 76.7 & 60.0 & 43.3 & 56.7 & 46.7 & 43.3 & 66.7 & 53.3 & 43.3 & 54.4 \\
KeyDiff & 76.7 & 63.3 & 50.0 & 66.7 & 60.0 & 60.0 & 71.7 & 61.7 & 55.0 & 62.8 \\
TriAttention & \underline{90.0} & \underline{90.0} & 76.7 & \underline{86.7} & 80.0 & \underline{66.7} & \underline{88.3} & 85.0 & 71.7 & 81.7 \\
KVzap & \textbf{96.7} & 86.7 & \underline{86.7} & \textbf{90.0} & \textbf{90.0} & \underline{66.7} & \textbf{93.3} & \underline{88.3} & \underline{76.7} & \underline{86.1} \\
\midrule
\rowcolor{GSrow}\method & \textbf{96.7} & \textbf{96.7} & \textbf{90.0} & \textbf{90.0} & \underline{86.7} & \textbf{83.3} & \textbf{93.3} & \textbf{91.7} & \textbf{86.7} & \textbf{90.6} \\
\bottomrule
\end{tabular*}

\caption{Decoding eviction accuracy (\%) on AIME24/25 with Qwen3.8-27B; columns: mean cache cap (entries per head). \underline{Underlined}: second best.}
\label{tab:aime}
\end{table*}

\subsection{32K Reasoning Decoding Transfer}
\label{sec:decode}

Trained solely on 1K-token generic excerpts ($\le 128$-token rollouts, zero positional features), \method\ transfers unchanged to online 32K \texttt{<think>} decoding on AIME24/25 (Table~\ref{tab:aime}, where two-pass KVzip cannot run every 128 steps), with all methods protecting 4 sink and 128 recent tokens (32 for SnapKV). \method\ matches or leads every baseline at all three caps ($93.3\%$, $91.7\%$, and $86.7\%$ at 4K, 3K, and 2K; $90.6\%$ mean vs.\ $95.0\%$ FullKV), outperforming KVzap ($86.1\%$ mean; $+10.0$ across all 60 problems and $+16.7$ on AIME25 at 2K), TriAttention ($81.7\%$), and the remaining training-free heuristics, which average $54.4\%$--$62.8\%$.

\section{Related Work}
\label{sec:related}

\paragraph{KV Importance Scoring.}
Query-aware policies accumulate observed attention via heavy hitters in H$_2$O~\citep{zhang2023h2o}, persistence in Scissorhands~\citep{liu2023scissorhands}, head profiling in FastGen~\citep{ge2024fastgen}, prompt tails in SnapKV~\citep{li2024snapkv}, or token redundancy in R-KV~\citep{cai2025rkv}, alongside sink-and-window retention in StreamingLLM~\citep{xiao2024streamingllm}. Query-agnostic heuristics~\citep{li2025scbench} score key $L_2$ norms~\citep{devoto2024l2norm}, mean-key dissimilarity in KeyDiff~\citep{park2025keydiff}, or calibrated RoPE statistics in TriAttention~\citep{mao2026triattention}, whereas learned methods probe prefill importance via lookahead tokens and LoRA in LookaheadKV~\citep{ahn2026lookaheadkv} or supervise prompt copying in two-pass KVzip~\citep{kim2025kvzip} and KVzap~\citep{jegou2026kvzap}. \method\ instead predicts continuation utility from native cross-memory recurrent signals, using one estimator across both prefill and online decoding eviction.

\paragraph{Cross-Layer and Cross-Head Allocation.}
Decaying schedules in PyramidKV~\citep{cai2024pyramidkv} overlook gate heterogeneity, while unbounded global selection in KVzip and KVzap imposes no per-layer floor. Head-wise quotas in DuoAttention~\citep{xiao2024duoattention}, AdaKV~\citep{feng2025adakv}, and HeadKV~\citep{fu2025headkv} create ragged head lengths incompatible with per-layer PagedAttention block tables~\citep{kwon2023vllm}; \method\ derives bounded gate-proportional layer budgets with head-uniform entry counts within each layer.

\paragraph{Hybrid Recurrent--Attention LLMs.}
Modern hybrids interleave selective SSMs~\citep{gu2023mamba,dao2024transformers}, RG-LRU~\citep{de2024griffin}, or gated delta recurrences~\citep{schlag2021linear,yang2024parallelizing,yang2025gated} with a few attention layers in Mamba~\citep{lieber2024jamba,ren2025samba,blakeman2025nemotronh} and linear/\gdn\ families~\citep{minimax2025minimax01,gu2025jetnemotron,kimi2025linear,merrill2026olmohybrid,qwen2026qwen38,qwen2026qwen38_27b,qiu2026qwen38next}. Prior work targets training-time layer pruning~\citep{blakeman2025nemotronh,gu2025jetnemotron}, whereas \method\ couples recurrent and attention memory at inference time.

\raggedbottom
\section{Conclusion}
\label{sec:conclusion}

We introduce \method, a query-agnostic KV cache eviction framework for hybrid LLMs that predicts attention token utility directly from native recurrent signals without extra forward passes. Keeping the backbone frozen, \method\ pairs lightweight, position-free per-head MLPs over preceding \gdn\ signals with bounded gate-proportional layer budgets aligned to PagedAttention block tables. Trained once on short generic excerpts to match leave-one-out continuation utility, \method\ outperforms training-free heuristics and costly prompt-reconstruction methods across both prefill and online decoding, especially in low-budget regimes, while substantially increasing long-context decoding throughput. It introduces less than $0.02\%$ additional parameters and incurs marginal scoring latency ($\le 1.4\%$ of prefill). Extending cross-memory eviction to broader hybrid families remains an interesting direction for future work.

\clearpage
\flushbottom
\section*{Limitations}
\label{sec:limitations}

\method\ targets hybrid architectures that interleave Gated DeltaNet (\gdn) layers with attention layers and reads the native write, decay, and value-norm signals $(\beta, g, \|v\|)$ of the recurrent layers; pure full-attention Transformers emit no such signals and are outside its scope. As analyzed in \S\ref{sec:prefill}, leave-one-out continuation utility aligns with semantic comprehension, question answering, and multi-step reasoning rather than verbatim copying of random strings, and reconstruction-based methods retain higher recall on the verbatim lookup tasks at moderate budgets.

\section*{Ethical Considerations}
This work studies inference-time memory efficiency for publicly released language models on standard public benchmarks (SCBench, SQuAD, GSM8K, AIME24, AIME25, RULER, and Nemotron pretraining excerpts) and introduces no human-subject data or safety-sensitive corpora. Because aggressive KV eviction ($p=0.1$) reduces exact verbatim string recall (\S\ref{sec:prefill}), deployments in high-stakes factual lookup settings should validate retention budgets against domain fidelity requirements. AI assistants were used solely for code and prose assistance; all experimental designs, measurements, and claims were verified by the authors.

\bibliography{references_verified}

\begin{thebibliography}{41}
\providecommand{\natexlab}[1]{#1}

\bibitem[{Ahn et~al.(2026)Ahn, Seong, Kedia, Kim, Jang, Lee, and Jeon}]{ahn2026lookaheadkv}
Jinwoo Ahn, Ingyu Seong, Akhil Kedia, Junhan Kim, Hyemi Jang, Kangwook Lee, and Yongkweon Jeon. 2026.
\newblock \href {https://openreview.net/forum?id=RVLMGPXt2i} {{LookaheadKV}: Fast and accurate {KV} cache eviction by glimpsing into the future without generation}.
\newblock In \emph{International Conference on Learning Representations (ICLR)}.
\newblock {arXiv}:2603.10899.

\bibitem[{Ainslie et~al.(2023)Ainslie, Lee-Thorp, de~Jong, Zemlyanskiy, Lebr{\'o}n, and Sanghai}]{ainslie2023gqa}
Joshua Ainslie, James Lee-Thorp, Michiel de~Jong, Yury Zemlyanskiy, Federico Lebr{\'o}n, and Sumit Sanghai. 2023.
\newblock \href {https://doi.org/10.18653/v1/2023.emnlp-main.298} {{GQA}: Training generalized multi-query transformer models from multi-head checkpoints}.
\newblock In \emph{Conference on Empirical Methods in Natural Language Processing (EMNLP)}, pages 4895--4901.
\newblock {arXiv}:2305.13245.

\bibitem[{Cai et~al.(2025{\natexlab{a}})Cai, Xiao, Sun, Luo, Zhang, Wan, Li, Zhou, Chang, Gu, Dong, Anandkumar, Asi, and Hu}]{cai2025rkv}
Zefan Cai, Wen Xiao, Hanshi Sun, Cheng Luo, Yikai Zhang, Ke~Wan, Yucheng Li, Yeyang Zhou, Li-Wen Chang, Jiuxiang Gu, Zhen Dong, Anima Anandkumar, Abedelkadir Asi, and Junjie Hu. 2025{\natexlab{a}}.
\newblock \href {https://proceedings.neurips.cc/paper_files/paper/2025/hash/57e0358aed314784cad14a26e1eba642-Abstract-Conference.html} {{R-KV}: Redundancy-aware {KV} cache compression for reasoning models}.
\newblock In \emph{Advances in Neural Information Processing Systems (NeurIPS)}, volume~38, pages 60980--61005.
\newblock {arXiv}:2505.24133.

\bibitem[{Cai et~al.(2025{\natexlab{b}})Cai, Zhang, Gao, Liu, Li, Liu, Lu, Xiong, Dong, Hu, and Xiao}]{cai2024pyramidkv}
Zefan Cai, Yichi Zhang, Bofei Gao, Yuliang Liu, Yucheng Li, Tianyu Liu, Keming Lu, Wayne Xiong, Yue Dong, Junjie Hu, and Wen Xiao. 2025{\natexlab{b}}.
\newblock \href {https://openreview.net/forum?id=ayi7qezU87} {{PyramidKV}: Dynamic {KV} cache compression based on pyramidal information funneling}.
\newblock In \emph{Conference on Language Modeling (COLM)}.
\newblock {arXiv}:2406.02069.

\bibitem[{Cobbe et~al.(2021)Cobbe, Kosaraju, Bavarian, Chen, Jun, Kaiser, Plappert, Tworek, Hilton, Nakano, Hesse, and Schulman}]{cobbe2021gsm8k}
Karl Cobbe, Vineet Kosaraju, Mohammad Bavarian, Mark Chen, Heewoo Jun, Lukasz Kaiser, Matthias Plappert, Jerry Tworek, Jacob Hilton, Reiichiro Nakano, Christopher Hesse, and John Schulman. 2021.
\newblock \href {https://arxiv.org/abs/2110.14168} {Training verifiers to solve math word problems}.
\newblock \emph{arXiv preprint arXiv:2110.14168}.

\bibitem[{Dao and Gu(2024)}]{dao2024transformers}
Tri Dao and Albert Gu. 2024.
\newblock \href {https://proceedings.mlr.press/v235/dao24a.html} {Transformers are {SSMs}: Generalized models and efficient algorithms through structured state space duality}.
\newblock In \emph{International Conference on Machine Learning (ICML)}, volume 235 of \emph{PMLR}, pages 10041--10071.
\newblock {arXiv}:2405.21060.

\bibitem[{De et~al.(2024)De, Smith, Fernando, Botev, Cristian-Muraru, Gu et~al.}]{de2024griffin}
Soham De, Samuel~L. Smith, Anushan Fernando, Aleksandar Botev, George Cristian-Muraru, Albert Gu, et~al. 2024.
\newblock \href {https://arxiv.org/abs/2402.19427} {{Griffin}: Mixing gated linear recurrences with local attention for efficient language models}.
\newblock \emph{arXiv preprint arXiv:2402.19427}.

\bibitem[{Devoto et~al.(2024)Devoto, Zhao, Scardapane, and Minervini}]{devoto2024l2norm}
Alessio Devoto, Yu~Zhao, Simone Scardapane, and Pasquale Minervini. 2024.
\newblock \href {https://doi.org/10.18653/v1/2024.emnlp-main.1027} {A simple and effective {$L_2$} norm-based strategy for {KV} cache compression}.
\newblock In \emph{Conference on Empirical Methods in Natural Language Processing (EMNLP)}, pages 18476--18499.
\newblock {arXiv}:2406.11430.

\bibitem[{Feng et~al.(2025)Feng, Lv, Cao, Xie, and Zhou}]{feng2025adakv}
Yuan Feng, Junlin Lv, Yukun Cao, Xike Xie, and S.~Kevin Zhou. 2025.
\newblock \href {https://proceedings.neurips.cc/paper_files/paper/2025/hash/a40ff56daab9f4808b1e18350c8a11ce-Abstract-Conference.html} {{Ada-KV}: Optimizing {KV} cache eviction by adaptive budget allocation for efficient {LLM} inference}.
\newblock In \emph{Advances in Neural Information Processing Systems (NeurIPS)}, volume~38, pages 113152--113188.
\newblock {arXiv}:2407.11550.

\bibitem[{Fu et~al.(2025)Fu, Cai, Asi, Xiong, Dong, and Xiao}]{fu2025headkv}
Yu~Fu, Zefan Cai, Abedelkadir Asi, Wayne Xiong, Yue Dong, and Wen Xiao. 2025.
\newblock \href {https://openreview.net/forum?id=FJFVmeXusW} {Not all heads matter: A head-level {KV} cache compression method with integrated retrieval and reasoning}.
\newblock In \emph{International Conference on Learning Representations (ICLR)}.
\newblock {arXiv}:2410.19258.

\bibitem[{Ge et~al.(2024)Ge, Zhang, Liu, Zhang, Han, and Gao}]{ge2024fastgen}
Suyu Ge, Yunan Zhang, Liyuan Liu, Minjia Zhang, Jiawei Han, and Jianfeng Gao. 2024.
\newblock \href {https://openreview.net/forum?id=uNrFpDPMyo} {Model tells you what to discard: Adaptive {KV} cache compression for {LLMs}}.
\newblock In \emph{International Conference on Learning Representations (ICLR)}.
\newblock {arXiv}:2310.01801.

\bibitem[{Gu and Dao(2024)}]{gu2023mamba}
Albert Gu and Tri Dao. 2024.
\newblock \href {https://openreview.net/forum?id=tEYskw1VY2} {Mamba: Linear-time sequence modeling with selective state spaces}.
\newblock In \emph{Conference on Language Modeling (COLM)}.
\newblock {arXiv}:2312.00752.

\bibitem[{Gu et~al.(2025)Gu, Hu, Xi, Chen, Yang, Han, and Cai}]{gu2025jetnemotron}
Yuxian Gu, Qinghao Hu, Haocheng Xi, Junyu Chen, Shang Yang, Song Han, and Han Cai. 2025.
\newblock \href {https://proceedings.neurips.cc/paper_files/paper/2025/hash/4364fbe773a77fb373df5c7f1d4cbb53-Abstract-Conference.html} {{Jet-Nemotron}: Efficient language model with post neural architecture search}.
\newblock In \emph{Advances in Neural Information Processing Systems (NeurIPS)}, volume~38, pages 47191--47218.
\newblock {arXiv}:2508.15884.

\bibitem[{Hsieh et~al.(2024)Hsieh, Sun, Kriman, Acharya, Rekesh, Jia, and Ginsburg}]{hsieh2024ruler}
Cheng-Ping Hsieh, Simeng Sun, Samuel Kriman, Shantanu Acharya, Dima Rekesh, Fei Jia, and Boris Ginsburg. 2024.
\newblock \href {https://openreview.net/forum?id=kIoBbc76Sy} {{RULER}: What's the real context size of your long-context language models?}
\newblock In \emph{Conference on Language Modeling (COLM)}.
\newblock {arXiv}:2404.06654.

\bibitem[{J{\'e}gou and Jeblick(2026)}]{jegou2026kvzap}
Simon J{\'e}gou and Maximilian Jeblick. 2026.
\newblock \href {https://arxiv.org/abs/2601.07891} {{KVzap}: Fast, adaptive, and faithful {KV} cache pruning}.
\newblock \emph{arXiv preprint arXiv:2601.07891}.

\bibitem[{Kim et~al.(2025)Kim, Kim, Kwon, Lee, Yun, and Song}]{kim2025kvzip}
Jang-Hyun Kim, Jinuk Kim, Sangwoo Kwon, Jae~W. Lee, Sangdoo Yun, and Hyun~Oh Song. 2025.
\newblock \href {https://proceedings.neurips.cc/paper_files/paper/2025/hash/f4eaa4b8f2d08edb3f0af990d56134ea-Abstract-Conference.html} {{KVzip}: Query-agnostic {KV} cache compression with context reconstruction}.
\newblock In \emph{Advances in Neural Information Processing Systems (NeurIPS)}, volume~38, pages 167563--167591.
\newblock {arXiv}:2505.23416.

\bibitem[{{Kimi Team} et~al.(2025){Kimi Team}, Zhang, Lin, Yao, Hu, Meng et~al.}]{kimi2025linear}
{Kimi Team}, Yu~Zhang, Zongyu Lin, Xingcheng Yao, Jiaxi Hu, Fanqing Meng, et~al. 2025.
\newblock \href {https://arxiv.org/abs/2510.26692} {{Kimi Linear}: An expressive, efficient attention architecture}.
\newblock \emph{arXiv preprint arXiv:2510.26692}.

\bibitem[{Kwon et~al.(2023)Kwon, Li, Zhuang, Sheng, Zheng, Yu, Gonzalez, Zhang, and Stoica}]{kwon2023vllm}
Woosuk Kwon, Zhuohan Li, Siyuan Zhuang, Ying Sheng, Lianmin Zheng, Cody~Hao Yu, Joseph~E. Gonzalez, Hao Zhang, and Ion Stoica. 2023.
\newblock \href {https://doi.org/10.1145/3600006.3613165} {Efficient memory management for large language model serving with {PagedAttention}}.
\newblock In \emph{ACM SIGOPS Symposium on Operating Systems Principles (SOSP)}, pages 611--626.
\newblock {arXiv}:2309.06180.

\bibitem[{Li et~al.(2025)Li, Jiang, Wu, Luo, Ahn, Zhang, Abdi, Li, Gao, Yang, and Qiu}]{li2025scbench}
Yucheng Li, Huiqiang Jiang, Qianhui Wu, Xufang Luo, Surin Ahn, Chengruidong Zhang, Amir~H. Abdi, Dongsheng Li, Jianfeng Gao, Yuqing Yang, and Lili Qiu. 2025.
\newblock \href {https://openreview.net/forum?id=gkUyYcY1W9} {{SCBench}: A {KV} cache-centric analysis of long-context methods}.
\newblock In \emph{International Conference on Learning Representations (ICLR)}.
\newblock {arXiv}:2412.10319.

\bibitem[{Li et~al.(2024)Li, Huang, Yang, Venkitesh, Locatelli, Ye, Cai, Lewis, and Chen}]{li2024snapkv}
Yuhong Li, Yingbing Huang, Bowen Yang, Bharat Venkitesh, Acyr Locatelli, Hanchen Ye, Tianle Cai, Patrick Lewis, and Deming Chen. 2024.
\newblock \href {https://proceedings.neurips.cc/paper_files/paper/2024/hash/28ab418242603e0f7323e54185d19bde-Abstract-Conference.html} {{SnapKV}: {LLM} knows what you are looking for before generation}.
\newblock In \emph{Advances in Neural Information Processing Systems (NeurIPS)}, volume~37, pages 22947--22970.
\newblock {arXiv}:2404.14469.

\bibitem[{Lieber et~al.(2024)Lieber, Lenz, Bata, Cohen, Osin, Dalmedigos, Safahi, Meirom, Belinkov, Shalev-Shwartz, Abend, Alon, Asida, Bergman, Glozman, Gokhman, Manevich, Ratner, Rozen, Shwartz, Zusman, and Shoham}]{lieber2024jamba}
Opher Lieber, Barak Lenz, Hofit Bata, Gal Cohen, Jhonathan Osin, Itay Dalmedigos, Erez Safahi, Shaked Meirom, Yonatan Belinkov, Shai Shalev-Shwartz, Omri Abend, Raz Alon, Tomer Asida, Amir Bergman, Roman Glozman, Michael Gokhman, Avashalom Manevich, Nir Ratner, Noam Rozen, and 3 others. 2024.
\newblock \href {https://arxiv.org/abs/2403.19887} {Jamba: A hybrid {Transformer-Mamba} language model}.
\newblock \emph{arXiv preprint arXiv:2403.19887}.

\bibitem[{Liu et~al.(2023)Liu, Desai, Liao, Wang, Xie, Xu, Kyrillidis, and Shrivastava}]{liu2023scissorhands}
Zichang Liu, Aditya Desai, Fangshuo Liao, Weitao Wang, Victor Xie, Zhaozhuo Xu, Anastasios Kyrillidis, and Anshumali Shrivastava. 2023.
\newblock \href {https://proceedings.neurips.cc/paper_files/paper/2023/hash/a452a7c6c463e4ae8fbdc614c6e983e6-Abstract-Conference.html} {Scissorhands: Exploiting the persistence of importance hypothesis for {LLM} {KV} cache compression at test time}.
\newblock In \emph{Advances in Neural Information Processing Systems (NeurIPS)}, volume~36, pages 52342--52364.
\newblock {arXiv}:2305.17118.

\bibitem[{Mao et~al.(2026)Mao, Lin, Huang, Xie, Fu, Zhuang, Han, and Chen}]{mao2026triattention}
Weian Mao, Xi~Lin, Wei Huang, Yuxin Xie, Tianfu Fu, Bohan Zhuang, Song Han, and Yukang Chen. 2026.
\newblock \href {https://proceedings.mlr.press/v306/mao26e.html} {{TriAttention}: Efficient long reasoning with trigonometric {KV} compression}.
\newblock In \emph{International Conference on Machine Learning (ICML)}, volume 306 of \emph{PMLR}, pages 86458--86477.
\newblock {arXiv}:2604.04921.

\bibitem[{Merrill et~al.(2026)Merrill, Li, Romero et~al.}]{merrill2026olmohybrid}
William Merrill, Yanhong Li, Tyler Romero, et~al. 2026.
\newblock \href {https://openreview.net/forum?id=LGhKvHMtVK} {{Olmo Hybrid}: From theory to practice and back}.
\newblock In \emph{Conference on Language Modeling (COLM)}.
\newblock {arXiv}:2604.03444.

\bibitem[{{MiniMax} et~al.(2025){MiniMax}, Li, Gong, Yang, Shan, Liu et~al.}]{minimax2025minimax01}
{MiniMax}, Aonian Li, Bangwei Gong, Bo~Yang, Boji Shan, Chang Liu, et~al. 2025.
\newblock \href {https://arxiv.org/abs/2501.08313} {{MiniMax-01}: Scaling foundation models with lightning attention}.
\newblock \emph{arXiv preprint arXiv:2501.08313}.

\bibitem[{{NVIDIA}(2025{\natexlab{a}})}]{blakeman2025nemotronh}
{NVIDIA}. 2025{\natexlab{a}}.
\newblock \href {https://arxiv.org/abs/2504.03624} {Nemotron-{H}: A family of accurate and efficient hybrid {Mamba-Transformer} models}.
\newblock \emph{arXiv preprint arXiv:2504.03624}.

\bibitem[{{NVIDIA}(2025{\natexlab{b}})}]{nvidia2025nemotronsample}
{NVIDIA}. 2025{\natexlab{b}}.
\newblock \href {https://huggingface.co/datasets/nvidia/Nemotron-Pretraining-Dataset-sample} {{Nemotron-Pretraining-Dataset-sample}}.
\newblock Hugging Face dataset.

\bibitem[{Park et~al.(2025)Park, Jones, Morse, Goel, Lee, and Lott}]{park2025keydiff}
Junyoung Park, Dalton Jones, Matthew~J. Morse, Raghavv Goel, Mingu Lee, and Christopher Lott. 2025.
\newblock \href {https://proceedings.neurips.cc/paper_files/paper/2025/hash/0907335ecf28faf15be54485dbcbe70e-Abstract-Conference.html} {{KeyDiff}: Key similarity-based {KV} cache eviction for long-context {LLM} inference in resource-constrained environments}.
\newblock In \emph{Advances in Neural Information Processing Systems (NeurIPS)}, volume~38, pages 5983--6019.
\newblock {arXiv}:2504.15364.

\bibitem[{Qiu et~al.(2026)Qiu, Wang, Li et~al.}]{qiu2026qwen38next}
Zihan Qiu, Zekun Wang, Xiao Li, et~al. 2026.
\newblock \href {https://arxiv.org/abs/2608.30320} {On the design of {Qwen3.8-Next} architecture: Evaluation, efficiency, and training stability}.
\newblock \emph{arXiv preprint arXiv:2608.30320}.

\bibitem[{Qiu et~al.(2025)Qiu, Wang, Zheng, Huang, Wen, Yang, Men, Yu, Huang, Huang, Liu, Zhou, and Lin}]{qiu2025gated}
Zihan Qiu, Zekun Wang, Bo~Zheng, Zeyu Huang, Kaiyue Wen, Songlin Yang, Rui Men, Le~Yu, Fei Huang, Suozhi Huang, Dayiheng Liu, Jingren Zhou, and Junyang Lin. 2025.
\newblock \href {https://doi.org/10.52202/085713-3345} {Gated attention for large language models: Non-linearity, sparsity, and attention-sink-free}.
\newblock In \emph{Advances in Neural Information Processing Systems (NeurIPS)}, volume~38, pages 110931--110957.
\newblock {arXiv}:2505.06708.

\bibitem[{{Qwen Team}(2026{\natexlab{a}})}]{qwen2026qwen38_27b}
{Qwen Team}. 2026{\natexlab{a}}.
\newblock \href {https://huggingface.co/Qwen/Qwen3.8-27B} {{Qwen3.8-27B}}.
\newblock Model card and configuration.

\bibitem[{{Qwen Team}(2026{\natexlab{b}})}]{qwen2026qwen38}
{Qwen Team}. 2026{\natexlab{b}}.
\newblock \href {https://qwen.ai/blog?id=qwen3.8} {{Qwen3.8-Max}: A new bar for coding and cowork}.
\newblock Blog post.

\bibitem[{Rajpurkar et~al.(2016)Rajpurkar, Zhang, Lopyrev, and Liang}]{rajpurkar2016squad}
Pranav Rajpurkar, Jian Zhang, Konstantin Lopyrev, and Percy Liang. 2016.
\newblock \href {https://doi.org/10.18653/v1/D16-1264} {{SQuAD}: 100,000+ questions for machine comprehension of text}.
\newblock In \emph{Proceedings of the 2016 Conference on Empirical Methods in Natural Language Processing (EMNLP)}, pages 2383--2392.

\bibitem[{Ren et~al.(2025)Ren, Liu, Lu, Shen, Liang, and Chen}]{ren2025samba}
Liliang Ren, Yang Liu, Yadong Lu, Yelong Shen, Chen Liang, and Weizhu Chen. 2025.
\newblock \href {https://openreview.net/forum?id=bIlnpVM4bc} {Samba: Simple hybrid state space models for efficient unlimited context language modeling}.
\newblock In \emph{International Conference on Learning Representations (ICLR)}.
\newblock {arXiv}:2406.07522.

\bibitem[{Schlag et~al.(2021)Schlag, Irie, and Schmidhuber}]{schlag2021linear}
Imanol Schlag, Kazuki Irie, and J{\"u}rgen Schmidhuber. 2021.
\newblock \href {https://proceedings.mlr.press/v139/schlag21a.html} {Linear transformers are secretly fast weight programmers}.
\newblock In \emph{International Conference on Machine Learning (ICML)}, volume 139 of \emph{PMLR}, pages 9355--9366.
\newblock {arXiv}:2102.11174.

\bibitem[{Shapley(1953)}]{shapley1953value}
Lloyd~S. Shapley. 1953.
\newblock \href {https://doi.org/10.1515/9781400881970-018} {A value for $n$-person games}.
\newblock In Harold~W. Kuhn and Albert~W. Tucker, editors, \emph{Contributions to the Theory of Games, Volume {II}}, volume~28 of \emph{Annals of Mathematics Studies}, pages 307--317. Princeton University Press.

\bibitem[{Xiao et~al.(2025)Xiao, Tang, Zuo, Guo, Yang, Tang, Fu, and Han}]{xiao2024duoattention}
Guangxuan Xiao, Jiaming Tang, Jingwei Zuo, Junxian Guo, Shang Yang, Haotian Tang, Yao Fu, and Song Han. 2025.
\newblock \href {https://openreview.net/forum?id=cFu7ze7xUm} {{DuoAttention}: Efficient long-context {LLM} inference with retrieval and streaming heads}.
\newblock In \emph{International Conference on Learning Representations (ICLR)}.
\newblock {arXiv}:2410.10819.

\bibitem[{Xiao et~al.(2024)Xiao, Tian, Chen, Han, and Lewis}]{xiao2024streamingllm}
Guangxuan Xiao, Yuandong Tian, Beidi Chen, Song Han, and Mike Lewis. 2024.
\newblock \href {https://openreview.net/forum?id=NG7sS51zVF} {Efficient streaming language models with attention sinks}.
\newblock In \emph{International Conference on Learning Representations (ICLR)}.
\newblock {arXiv}:2309.17453.

\bibitem[{Yang et~al.(2025)Yang, Kautz, and Hatamizadeh}]{yang2025gated}
Songlin Yang, Jan Kautz, and Ali Hatamizadeh. 2025.
\newblock \href {https://openreview.net/forum?id=r8H7xhYPwz} {Gated delta networks: Improving {Mamba2} with delta rule}.
\newblock In \emph{International Conference on Learning Representations (ICLR)}.
\newblock {arXiv}:2412.06464.

\bibitem[{Yang et~al.(2024)Yang, Wang, Zhang, Shen, and Kim}]{yang2024parallelizing}
Songlin Yang, Bailin Wang, Yu~Zhang, Yikang Shen, and Yoon Kim. 2024.
\newblock \href {https://proceedings.neurips.cc/paper_files/paper/2024/hash/d13a3eae72366e61dfdc7eea82eeb685-Abstract-Conference.html} {Parallelizing linear transformers with the delta rule over sequence length}.
\newblock In \emph{Advances in Neural Information Processing Systems (NeurIPS)}, volume~37, pages 115491--115522.
\newblock {arXiv}:2406.06484.

\bibitem[{Zhang et~al.(2023)Zhang, Sheng, Zhou, Chen, Zheng, Cai, Song, Tian, R{\'e}, Barrett, Wang, and Chen}]{zhang2023h2o}
Zhenyu Zhang, Ying Sheng, Tianyi Zhou, Tianlong Chen, Lianmin Zheng, Ruisi Cai, Zhao Song, Yuandong Tian, Christopher R{\'e}, Clark Barrett, Zhangyang Wang, and Beidi Chen. 2023.
\newblock \href {https://proceedings.neurips.cc/paper_files/paper/2023/hash/6ceefa7b15572587b78ecfcebb2827f8-Abstract-Conference.html} {{H$_2$O}: Heavy-hitter oracle for efficient generative inference of large language models}.
\newblock In \emph{Advances in Neural Information Processing Systems (NeurIPS)}, volume~36, pages 34661--34710.
\newblock {arXiv}:2306.14048.

\end{thebibliography}

\clearpage
\appendix
\raggedbottom
\widowpenalty=10000 \clubpenalty=10000   %
\let\GSunderscore\_ \renewcommand{\_}{\GSunderscore\allowbreak}   %
\renewcommand{\dblfloatpagefraction}{0.90}\renewcommand{\floatpagefraction}{0.95}
\renewcommand{\dbltopfraction}{0.85}\setcounter{dbltopnumber}{3}
\renewcommand{\topfraction}{0.90}\renewcommand{\textfraction}{0.08}\setcounter{topnumber}{4}\setcounter{totalnumber}{5}
\newlength{\GSgroupsep}\setlength{\GSgroupsep}{14pt}
\setlength{\dbltextfloatsep}{12pt plus 2pt minus 2pt}
\setlength{\dblfloatsep}{10pt plus 2pt minus 2pt}
\makeatletter
\setlength{\@dblfptop}{0pt}\setlength{\@dblfpsep}{22pt}\setlength{\@dblfpbot}{0pt plus 1fil}
\setlength{\@fptop}{0pt}\setlength{\@fpsep}{22pt}\setlength{\@fpbot}{0pt plus 1fil}
\makeatother
\section{Evaluation Protocol}
\label{app:protocol}

All prefill eviction methods share one deployment protocol (baseline scoring details in Appendix~\ref{app:details}; QSA block pooling in Appendix~\ref{app:flashnext}):
(1)~Prefill the complete shared prefix once into an unpruned parent session (middle-truncating above 262K per SCBench), setting $n_{\mathrm{prot}} \le 256$ to the chat/instruction header before the context (4-token-aligned on QSA).
(2)~Compute each method's query-agnostic prefix scores on that parent.
(3)~Create one physically compacted child session at a time via \texttt{fork\_session}, verifying physical page release.
(4)~Reuse the child cache across all follow-up queries of that context (restoring recurrent states and scheduling only query tokens, token-identical to a fresh run).
Prefix caching is disabled, \texttt{max\_num\_seqs=1}, and SHA-256 over token IDs rejects mismatched or preempted sessions.

\paragraph{Benchmark and Scoring Rules.}
On SCBench, the composite tasks \texttt{summary\_with\_needles} and \texttt{repoqa\_and\_kv} enter the 12-task mean via the unweighted mean of their two sub-tasks. RULER-4096 uses 100 official items per task (seed 42) across eight needle-in-a-haystack variants, variable tracking, word extraction, and two QA tasks. Dataset files for SQuAD and GSM8K are pinned by SHA-256, with protocols adapted from KVzip~\citep{kim2025kvzip}:
(1)~SQuAD v1.1 uses the first 100 unique passages of \texttt{train-v1.1.json} (\texttt{max\_new\_tokens=64}), scored by KVzip normalized answer-inclusion match (lowercased; punctuation, articles, and whitespace stripped; \texttt{zero}--\texttt{nine} mapped to digits).
(2)~GSM8K uses the first 100 test items whose setup sentences contain $\ge 72$ lexical tokens (\texttt{max\_new\_tokens=1,024}), scored by exact numeric match parsing \texttt{The answer is \dots} (\texttt{\#\#\#\#} and \texttt{\textbackslash boxed\{\}} fallback).
(3)~AIME24/25 sample one trajectory per problem (\texttt{seed + index}, \texttt{enable\_thinking=True}, \texttt{max\_length=32,768}); if \texttt{</think>} has not appeared by step $30{,}720$, a forced \texttt{</think>} prompt elicits the final answer, scored by exact integer match in post-\texttt{</think>} \texttt{\textbackslash boxed\{\}} (official AIME~I/II problems).

\paragraph{SCBench Parser Fixes.}
Applied identically across all methods, budgets, and models (fixed prior to inspecting rankings):
(1)~\texttt{qa\_chn}: the official script iterates over characters of a single reference string rather than a reference list; we normalize predictions and full references and take max character-level $F_1$ ($+5.6$ points across 8,694 rows on 27B; $+6.0$ on FullKV).
(2)~\texttt{choice\_eng}: the official script reads the first output character as the option letter, mis-scoring \texttt{Based on \dots} as \texttt{B}; we accept only explicit option formats (\texttt{B}, \texttt{B. Beta}, \texttt{(B)}, \texttt{Option B}, \texttt{Answer: B}), lowering the mean by $-12.3$ points ($-11.4$ on FullKV).
With the unmodified scorer, the 27B 12-task mean shifts by $+0.3$--$1.4$ points with zero ranking change.

\section{Utility Label and Shapley Validation}
\label{app:label}

\begin{table}[t]
\centering
\footnotesize
\setlength{\tabcolsep}{2.8pt}
\begin{tabular*}{\linewidth}{@{\extracolsep{\fill}}lccc}
\toprule
Score vs.\ $\phi_{\mathrm{mean}}$ & $\rho$ & Ov@10 & Cov@10 \\
\midrule
\rowcolor{GSrow}$\U_{\max}$ (deployed label) & $+0.901$ & 69.2 & 88.2 \\
$\U_{\mathrm{mean}}$ & $+0.950$ & 83.8 & 95.5 \\
Attention max & $+0.909$ & 70.3 & 89.1 \\
Attention mean & $+0.948$ & 82.2 & 95.2 \\
\midrule
Value norm $\|v\|$ & $+0.202$ & 20.4 & 32.6 \\
\method\ MLP & $+0.487$ & 38.5 & 57.2 \\
Random & $-0.003$ & 9.8 & 18.5 \\
\bottomrule
\end{tabular*}

\caption{Agreement with multi-entry Shapley values $\phi_{\mathrm{mean}}$ over 768 head games on Qwen3.8-27B (Ov@10/Cov@10 in \%).}
\label{tab:shapley}
\end{table}

Each training document is prefilled once, and its $M=4$ continuations (\S\ref{sec:background}) are sampled at $T=1.0$, top-$p=0.95$, and top-$k=20$. Causal attention from continuation tokens over the context is recomputed in FP32, and Eq.~\ref{eq:utility} is evaluated in closed form via cross-head inner products $W_O^{(j)\top}W_O^{(j')}$ ($j,j'\in\mathcal{Q}(h)$). Because $v_i^{\top}W_O^{(j)\top}W_O^{(j')} v_i$ depends on entry $i$ alone once the output gate is omitted, perturbations run in one pass per entry (verified on every document against native forward outputs and brute-force key deletion).

\begin{figure*}[t]
\centering
\includegraphics[width=0.75\textwidth]{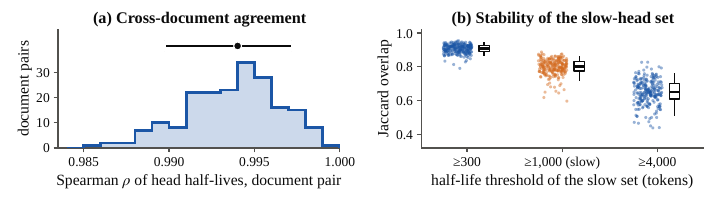}
\caption{Cross-document stability of \gdn\ head half-lives on Qwen3.8-27B. \textbf{(a)}~Spearman $\rho$ of head half-lives across random document pairs (dot and bar: median and 10--90th percentiles). \textbf{(b)}~Jaccard overlap of per-document slow-head sets with the global set (box: quartiles; whiskers: 5--95th percentiles).}
\label{fig:obs1}
\end{figure*}

\begin{figure*}[t]
\centering
\includegraphics[width=0.75\textwidth]{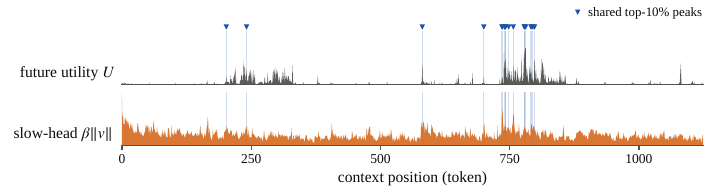}
\caption{Token-level alignment of slow-head $\beta\|v\|$ and future utility $\U$ at attention layer~55 on a median-correlation document ($\blacktriangledown$: tokens in the top 10\% of both).}
\label{fig:obs2example}
\end{figure*}

\paragraph{Batch-Eviction Identity and Shapley Validation.}
Writing $\Delta r_j(t,i)=[a_j(t,i)/(1-a_j(t,i))]W_O^{(j)}(v_i-o_j(t))$, simultaneously evicting $E=\{1,\dots,n\}\setminus S$ with head mass $A_j(t,E)=\sum_{i\in E}a_j(t,i)<1$ gives exact softmax-renormalized shift
\begin{align*}
\Delta r(t,E) = \!\!\sum_{j\in\mathcal{Q}(h)}\!\! &\frac{1}{1-A_j(t,E)} \\
&\times \sum_{i\in E}\big(1-a_j(t,i)\big)\Delta r_j(t,i).
\end{align*}
Entrywise $a_j(t,i)\ll 1$ alone does not bound $\|\Delta r(t,E)\|$ via $\sum_{i\in E}\U(i)$: $A_j(t,E)$ may still approach $1$, and the head-specific factor $1/(1-A_j(t,E))$ breaks cross-head cancellations in $\Delta r(t,i)=\sum_j\Delta r_j(t,i)$. An upper bound requires $A_j(t,E)\le\alpha<1$ and per-head norms,
\begin{equation*}
\|\Delta r(t,E)\|\le\tfrac{1}{1-\alpha}\sum_{i\in E,j}(1-a_j(t,i))\|\Delta r_j(t,i)\|;
\end{equation*}
for $A_j(t,E)\ll 1$, $(1-a_j(t,i))/(1-A_j(t,E))=1+O(A_j(t,E))$ makes $\sum_{i\in E}\Delta r(t,i)$ a local first-order proxy. To test multi-entry interactions directly, Table~\ref{tab:shapley} evaluates sampled Shapley values~\citep{shapley1953value} on 12 Nemotron 512-token documents on Qwen3.8-27B ($768$ head games, $2{,}048$ permutations/head; split-half $\rho$ median $0.9995$, min $0.9921$) for $v_t(S)=-\|r_t(S)-r_t(N)\|_2$ averaged over continuation steps ($\phi_{\mathrm{mean}}$). Step-mean scores ($\U_{\mathrm{mean}}$, Attention mean) align more closely with $\phi_{\mathrm{mean}}$ ($95.5\%$, $95.2\%$ Cov@10), whereas against $\max_{t\in\mathcal{T}}\phi_t(i)$, $\U_{\max}$ and Attention max reach $\rho=+0.946$/$+0.949$ and $95.4\%$/$95.5\%$ Cov@10 (vs.\ $86.7\%$ for $\U_{\mathrm{mean}}$). This short-context check tests coalition interactions rather than ranking $\U_{\max}$ above continuation attention: under joint top-$p$ eviction ($p\in\{0.5,0.3,0.1\}$), $\U_{\max}$ ($0.036$, $0.081$, $0.215$) and Attention max ($0.036$, $0.081$, $0.213$) match $\phi_{\mathrm{mean}}$ ($0.039$, $0.080$, $0.204$), while \method\ MLPs cut random error by $27\%$--$50\%$ ($0.095$, $0.167$, $0.312$ vs.\ $0.190$, $0.278$, $0.427$).

\section{Details Behind the Empirical Observations}
\label{app:obs}

\begin{figure}[t]
\centering
\includegraphics[width=0.85\linewidth]{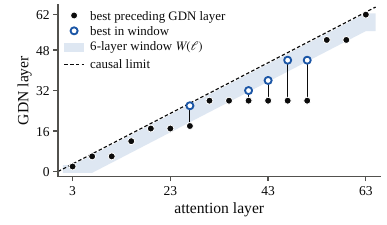}
\caption{Most predictive preceding \gdn\ layer per attention layer vs.\ the six-layer window $W(\ell)$.}
\label{fig:obs2layers}
\end{figure}

\begin{table*}[t]
\begin{minipage}{\textwidth}\centering\captionsetup{type=table}
{\footnotesize\renewcommand{\arraystretch}{\GSrowstretch}
\setlength{\tabcolsep}{2.6pt}
\begin{tabular*}{\linewidth}{@{\extracolsep{\fill}}lccccccc}
\toprule
\multirow{2}{*}{Predictor} & \multirow{2}{*}{$\rho$ ($\uparrow$)} & \multicolumn{3}{c}{Ov@$k$ ($\uparrow$)} & \multicolumn{3}{c}{Cov@$k$ ($\uparrow$)} \\
\cmidrule(lr){3-5}\cmidrule(lr){6-8}
 & & 5 & 10 & 20 & 5 & 10 & 20 \\
\midrule
\multicolumn{8}{@{}l}{\textit{Untrained signals}} \\[1pt]
Random & $+0.000$ & 5.0 & 10.0 & 19.9 & 14.1 & 19.0 & 27.8 \\
Age (older first) & $+0.071$ & 11.6 & 19.2 & 29.4 & 24.4 & 31.2 & 40.0 \\
$\|v\|$, preceding \gdn\ layer & $+0.121$ & 7.9 & 13.9 & 24.8 & 19.5 & 24.8 & 34.1 \\
$\beta$, window mean & $+0.107$ & 8.9 & 14.9 & 25.4 & 21.2 & 26.0 & 34.6 \\
$-g$, window mean & $+0.106$ & 6.9 & 12.9 & 24.2 & 18.7 & 23.6 & 32.7 \\
Retention $\exp(g)$, window mean & $-0.116$ & 3.7 & 7.9 & 16.9 & 11.5 & 16.0 & 24.0 \\
$\beta\|v\|$, window mean & $+0.103$ & 8.5 & 14.4 & 24.7 & 20.4 & 25.2 & 33.9 \\
$\|v\|$ into slow heads & $+0.126$ & 9.0 & 15.3 & 25.8 & 21.1 & 26.4 & 35.3 \\
Slow-head $\beta\|v\|$ & $+0.229$ & 10.6 & 18.2 & 30.3 & 24.3 & 30.6 & 40.9 \\
Best single \gdn\ head (oracle$^*$) & $+0.232$ & 15.2 & 22.6 & 33.1 & 30.2 & 36.0 & 45.4 \\
\midrule
\multicolumn{8}{@{}l}{\textit{Learned predictors (leave-one-corpus-out)}} \\[1pt]
Per-head MLP, 3 age features (control) & $+0.149$ & 11.1 & 17.9 & 27.9 & 23.6 & 29.6 & 38.1 \\
Ridge regression, 864 \gdn\ features & {\boldmath$+0.527$} & 27.0 & 36.4 & 47.9 & 46.1 & 52.6 & 62.3 \\
\rowcolor{GSrow}Per-head MLP, 864 \gdn\ features (\method) & $+0.523$ & \textbf{29.8} & \textbf{38.1} & \textbf{48.4} & \textbf{48.5} & \textbf{54.6} & \textbf{64.0} \\
\bottomrule
\end{tabular*}
\par}
\caption{Full version of Table~\ref{tab:obs3}: Ov@$k$/Cov@$k$ (\%) at $k\in\{5,10,20\}$ for all predictors ($^*$: selected in-sample).}
\label{tab:obs3full}
\end{minipage}
\par\vspace{\GSgroupsep}
\begin{minipage}{\textwidth}\centering\captionsetup{type=figure}
\includegraphics[width=\textwidth]{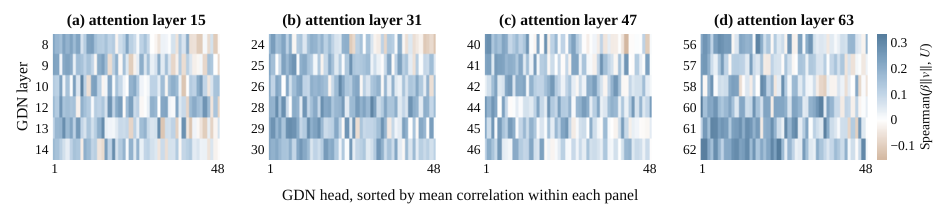}
\caption{Spearman $\rho$ between $\beta\|v\|$ and $\U$ for every (\gdn\ layer, head) cell in the six-layer windows $W(\ell)$ of four attention layers.}
\label{fig:obs3cells}
\end{minipage}
\end{table*}

\begin{table}[t]
\centering
\footnotesize
\setlength{\tabcolsep}{1.8pt}
\begin{tabular*}{\linewidth}{@{\extracolsep{\fill}}lcccc}
\toprule
Signal input & $\rho$ & Ov@10 & Cov@10 & Cov@20 \\
\midrule
KeyDiff & $+0.213$ & 18.1 & 30.3 & 40.3 \\
TriAttention & $+0.389$ & 28.8 & 44.0 & 54.2 \\
\rowcolor{GSrow}\gdn\ (\method) & {\boldmath$+0.516$} & \textbf{38.2} & \textbf{55.5} & \textbf{64.4} \\
\bottomrule
\end{tabular*}

\caption{Width-64 MLPs trained on intra-attention vs.\ \gdn\ inputs to predict $\U$ (LOCO, 240 documents; Ov/Cov in \%).}
\label{tab:signal_mlp}
\end{table}

\begin{figure*}[t]
\begin{minipage}{\textwidth}\centering\captionsetup{type=figure}
\includegraphics[width=0.75\textwidth]{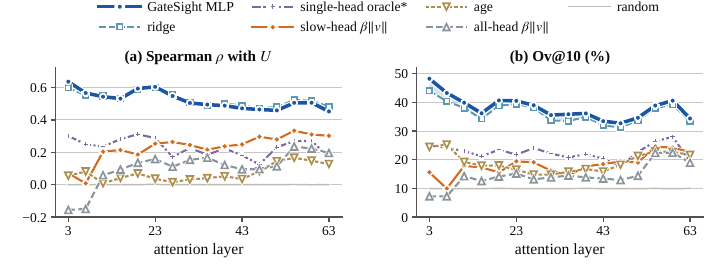}
\caption{Per-attention-layer \textbf{(a)}~Spearman $\rho$ and \textbf{(b)}~Ov@10 of the predictors in Table~\ref{tab:obs3full} (shaded: MLP gain over ridge; $^*$: selected in-sample).}
\label{fig:obs3layers}
\end{minipage}
\par\vspace{\GSgroupsep}
\begin{minipage}{\textwidth}\centering\captionsetup{type=table}
{\footnotesize\renewcommand{\arraystretch}{\GSrowstretch}
\setlength{\tabcolsep}{1.8pt}
\begin{tabular*}{\linewidth}{@{\extracolsep{\fill}}lcccccccccc}
\toprule
\multirow{2}{*}{Evaluation set} & \multicolumn{2}{c}{Random} & \multicolumn{2}{c}{Age} & \multicolumn{2}{c}{$\beta\|v\|$, window} & \multicolumn{2}{c}{Slow-head $\beta\|v\|$} & \multicolumn{2}{c}{\method} \\
\cmidrule(lr){2-3}\cmidrule(lr){4-5}\cmidrule(lr){6-7}\cmidrule(lr){8-9}\cmidrule(lr){10-11}
 & $\rho$ & Ov@10 & $\rho$ & Ov@10 & $\rho$ & Ov@10 & $\rho$ & Ov@10 & $\rho$ & Ov@10 \\
\midrule
Nemotron, 1K (240 docs) & $+0.00$ & 9.9 & $+0.09$ & 19.3 & $+0.11$ & 14.5 & $+0.23$ & 18.3 & {\boldmath$+0.59$} & \textbf{45.3} \\
Nemotron, 2K (40) & $+0.00$ & 10.0 & $+0.19$ & 29.8 & $+0.18$ & 19.0 & $+0.34$ & 26.3 & {\boldmath$+0.58$} & \textbf{43.8} \\
Nemotron, 4K (40) & $+0.00$ & 10.0 & $+0.10$ & 24.4 & $+0.16$ & 18.0 & $+0.34$ & 27.2 & {\boldmath$+0.56$} & \textbf{40.3} \\
Nemotron, 8K (19) & $+0.00$ & 9.9 & $+0.23$ & 31.9 & $+0.13$ & 17.9 & $+0.33$ & 27.8 & {\boldmath$+0.56$} & \textbf{40.7} \\
RULER-8K (60 contexts) & $-0.00$ & 10.0 & $+0.12$ & 24.3 & $+0.21$ & 25.3 & $+0.30$ & 33.2 & {\boldmath$+0.39$} & \textbf{39.4} \\
\bottomrule
\end{tabular*}
\par}
\caption{Zero-shot length and task transfer of the 1K-trained estimator on Qwen3.8-27B (Ov@10 in \%; only the 1K row is in-sample).}
\label{tab:transfer}
\end{minipage}
\par\vspace{\GSgroupsep}
\begin{minipage}{\textwidth}\centering\captionsetup{type=table}
{\footnotesize\renewcommand{\arraystretch}{\GSrowstretch}
\begin{tabular*}{\textwidth}{@{\extracolsep{\fill}}llcccccccccc}
\toprule
Model & Method & 0.9 & 0.8 & 0.7 & 0.6 & 0.5 & 0.4 & 0.3 & 0.2 & 0.1 & Mean \\
\midrule
\multirow{7}{*}{\begin{tabular}[c]{@{}l@{}}Qwen3.8-\\27B\end{tabular}} & FullKV & \multicolumn{10}{c}{65.0} \\
 & SnapKV & 60.0 & 55.3 & 51.4 & 47.5 & 43.6 & 40.8 & 37.3 & 34.2 & 30.4 & 44.5 \\
 & TriAttention & 62.7 & 59.9 & 56.4 & 52.6 & 48.8 & 45.6 & 42.4 & 38.6 & 32.5 & 48.8 \\
 & KeyDiff & 62.4 & 57.4 & 52.1 & 47.6 & 43.7 & 41.4 & 39.2 & 36.9 & 33.3 & 46.0 \\
 & KVzap & \textbf{64.8} & 64.5 & 63.8 & 63.5 & 61.5 & 58.5 & 51.6 & 41.7 & 27.2 & 55.2 \\
 & KVzip$^\dagger$ & \textbf{64.8} & \textbf{64.7} & \textbf{64.5} & \textbf{64.1} & \textbf{62.6} & \textbf{60.4} & \textbf{55.1} & \textbf{41.9} & 27.9 & \textbf{56.2} \\
\rowcolor{GSrow}\cellcolor{white} & \method & 64.5 & 64.1 & 62.8 & 60.2 & 57.4 & 52.8 & 45.0 & 40.9 & \textbf{37.1} & 53.9 \\
\midrule
\multirow{7}{*}{\begin{tabular}[c]{@{}l@{}}Qwen3.8-\\Flash-Next\end{tabular}} & FullKV & \multicolumn{10}{c}{64.9} \\
 & SnapKV & 57.9 & 53.3 & 48.9 & 45.6 & 41.1 & 37.6 & 33.7 & 29.0 & 23.3 & 41.2 \\
 & TriAttention & 63.4 & 59.2 & 54.3 & 49.4 & 45.8 & 43.0 & 40.3 & 37.7 & 33.2 & 47.4 \\
 & KeyDiff & 58.8 & 54.9 & 50.9 & 47.7 & 45.5 & 43.1 & 40.7 & 37.4 & 33.9 & 45.9 \\
 & KVzap & \textbf{64.7} & 64.2 & 62.8 & 59.9 & 55.8 & 46.2 & 31.3 & 23.8 & 21.6 & 47.8 \\
 & KVzip$^\dagger$ & 64.6 & \textbf{65.2} & \textbf{64.4} & \textbf{64.1} & \textbf{62.6} & \textbf{59.0} & \textbf{49.2} & 30.8 & 23.2 & \textbf{53.7} \\
\rowcolor{GSrow}\cellcolor{white} & \method & 63.0 & 60.5 & 58.2 & 50.8 & 47.0 & 44.6 & 42.5 & \textbf{40.7} & \textbf{38.4} & 49.5 \\
\bottomrule
\end{tabular*}
\par}
\caption{SCBench 12-task mean accuracy (\%) vs.\ kept fraction $p$ ($^\dagger$: two-pass).}
\label{tab:budgets_scbench}
\end{minipage}
\end{figure*}

\begin{table*}[t]
\begin{minipage}{\textwidth}\centering\captionsetup{type=table}
{\footnotesize\renewcommand{\arraystretch}{\GSrowstretch}
\begin{tabular*}{\textwidth}{@{\extracolsep{\fill}}llcccccccccc}
\toprule
Model & Method & 0.9 & 0.8 & 0.7 & 0.6 & 0.5 & 0.4 & 0.3 & 0.2 & 0.1 & Mean \\
\midrule
\multirow{7}{*}{\begin{tabular}[c]{@{}l@{}}Qwen3.8-\\27B\end{tabular}} & FullKV & \multicolumn{10}{c}{97.2} \\
 & SnapKV & 93.2 & 83.4 & 75.8 & 66.0 & 56.5 & 45.9 & 36.5 & 27.0 & 16.4 & 55.6 \\
 & TriAttention & \textbf{97.6} & 97.0 & 95.1 & 92.4 & 84.6 & 77.2 & 65.8 & 44.5 & 24.0 & 75.4 \\
 & KeyDiff & 96.5 & 93.0 & 88.1 & 80.5 & 68.9 & 57.1 & 44.4 & 28.4 & 13.1 & 63.3 \\
 & KVzap & 97.5 & 97.5 & \textbf{97.5} & \textbf{97.4} & 95.6 & 92.5 & 81.5 & 61.5 & 18.5 & 82.2 \\
 & KVzip$^\dagger$ & 97.1 & \textbf{97.6} & \textbf{97.5} & 96.9 & 95.6 & 92.4 & 82.1 & 53.4 & 18.0 & 81.2 \\
\rowcolor{GSrow}\cellcolor{white} & \method & 97.4 & 97.4 & \textbf{97.5} & 97.2 & \textbf{96.4} & \textbf{93.9} & \textbf{88.2} & \textbf{77.5} & \textbf{54.4} & \textbf{88.9} \\
\midrule
\multirow{7}{*}{\begin{tabular}[c]{@{}l@{}}Qwen3.8-\\Flash-Next\end{tabular}} & FullKV & \multicolumn{10}{c}{96.5} \\
 & SnapKV & 90.8 & 83.5 & 75.2 & 66.4 & 57.0 & 47.8 & 38.9 & 29.1 & 17.5 & 56.2 \\
 & TriAttention & 91.5 & 84.6 & 77.9 & 67.6 & 61.4 & 51.1 & 42.1 & 31.8 & 18.9 & 58.5 \\
 & KeyDiff & 92.2 & 84.8 & 75.2 & 69.0 & 57.2 & 47.6 & 37.2 & 23.9 & 14.2 & 55.7 \\
 & KVzap & 96.8 & 95.1 & 89.4 & 75.2 & 48.8 & 28.9 & 16.4 & 13.4 & 10.9 & 52.8 \\
 & KVzip$^\dagger$ & \textbf{97.0} & \textbf{96.2} & 91.4 & 70.9 & 42.0 & 25.8 & 17.8 & 13.1 & 9.9 & 51.6 \\
\rowcolor{GSrow}\cellcolor{white} & \method & 96.0 & 95.8 & \textbf{93.8} & \textbf{87.8} & \textbf{79.9} & \textbf{71.0} & \textbf{59.0} & \textbf{47.9} & \textbf{29.5} & \textbf{73.4} \\
\bottomrule
\end{tabular*}
\par}
\caption{SQuAD v1.1 accuracy (\%) vs.\ kept fraction $p$ ($^\dagger$: two-pass).}
\label{tab:budgets_squad}
\end{minipage}
\par\vspace{\GSgroupsep}
\begin{minipage}{\textwidth}\centering\captionsetup{type=table}
{\footnotesize\renewcommand{\arraystretch}{\GSrowstretch}
\begin{tabular*}{\textwidth}{@{\extracolsep{\fill}}llcccccccccc}
\toprule
Model & Method & 0.9 & 0.8 & 0.7 & 0.6 & 0.5 & 0.4 & 0.3 & 0.2 & 0.1 & Mean \\
\midrule
\multirow{7}{*}{\begin{tabular}[c]{@{}l@{}}Qwen3.8-\\27B\end{tabular}} & FullKV & \multicolumn{10}{c}{90.0} \\
 & SnapKV & 88.0 & 66.0 & 50.0 & 30.0 & 16.0 & 7.0 & 4.0 & 1.0 & 3.0 & 29.4 \\
 & TriAttention & 90.0 & \textbf{93.0} & \textbf{92.0} & 89.0 & 83.0 & 69.0 & 47.0 & 22.0 & 6.0 & 65.7 \\
 & KeyDiff & \textbf{92.0} & 89.0 & 88.0 & 81.0 & 61.0 & 37.0 & 24.0 & 8.0 & 4.0 & 53.8 \\
 & KVzap & 90.0 & 90.0 & \textbf{92.0} & 90.0 & 87.0 & 85.0 & 67.0 & 42.0 & 6.0 & 72.1 \\
 & KVzip$^\dagger$ & 88.0 & 88.0 & 91.0 & \textbf{91.0} & \textbf{92.0} & \textbf{91.0} & 75.0 & 27.0 & 5.0 & 72.0 \\
\rowcolor{GSrow}\cellcolor{white} & \method & 90.0 & 90.0 & \textbf{92.0} & 87.0 & \textbf{92.0} & 90.0 & \textbf{85.0} & \textbf{72.0} & \textbf{43.0} & \textbf{82.3} \\
\midrule
\multirow{7}{*}{\begin{tabular}[c]{@{}l@{}}Qwen3.8-\\Flash-Next\end{tabular}} & FullKV & \multicolumn{10}{c}{93.0} \\
 & SnapKV & 81.0 & 62.0 & 46.0 & 28.0 & 14.0 & 8.0 & 3.0 & 5.0 & 3.0 & 27.8 \\
 & TriAttention & 78.0 & 61.0 & 45.0 & 39.0 & 17.0 & 5.0 & 5.0 & 6.0 & 2.0 & 28.7 \\
 & KeyDiff & 89.0 & 72.0 & 49.0 & 32.0 & 21.0 & 15.0 & 10.0 & 3.0 & 1.0 & 32.4 \\
 & KVzap & \textbf{94.0} & 89.0 & 77.0 & 51.0 & 23.0 & 5.0 & 2.0 & 2.0 & 2.0 & 38.3 \\
 & KVzip$^\dagger$ & \textbf{94.0} & \textbf{95.0} & 87.0 & 56.0 & 27.0 & 8.0 & 1.0 & 6.0 & 2.0 & 41.8 \\
\rowcolor{GSrow}\cellcolor{white} & \method & 93.0 & 92.0 & \textbf{94.0} & \textbf{84.0} & \textbf{62.0} & \textbf{44.0} & \textbf{27.0} & \textbf{9.0} & \textbf{4.0} & \textbf{56.6} \\
\bottomrule
\end{tabular*}
\par}
\caption{GSM8K accuracy (\%) vs.\ kept fraction $p$ ($^\dagger$: two-pass).}
\label{tab:budgets_gsm8k}
\end{minipage}
\end{table*}

\begin{figure*}[t]
\begin{minipage}{\textwidth}\centering\captionsetup{type=figure}
\includegraphics[width=\textwidth]{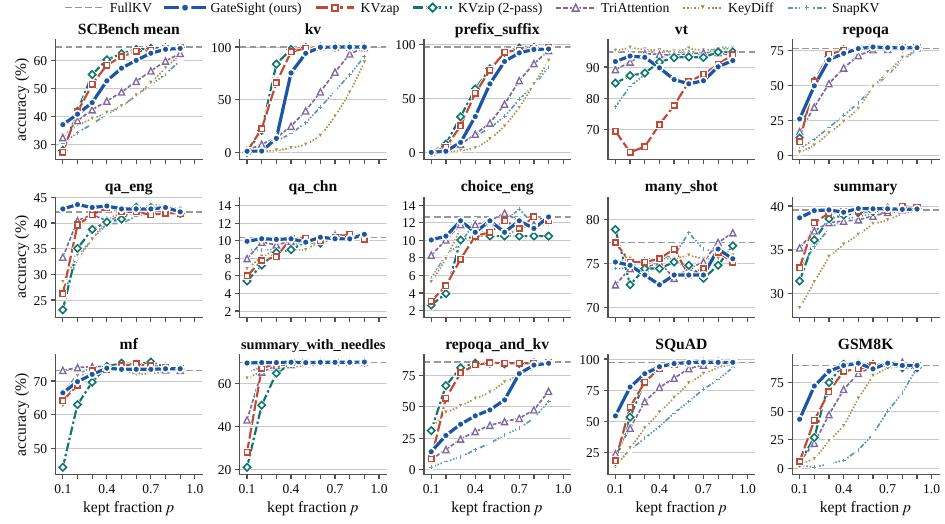}
\caption{Per-task accuracy (\%) vs.\ kept fraction $p$ on Qwen3.8-27B: SCBench, SQuAD v1.1, and GSM8K (dashed: FullKV).}
\label{fig:pertask}
\end{minipage}
\par\vspace{\GSgroupsep}
\begin{minipage}{\textwidth}\centering\captionsetup{type=figure}
\includegraphics[width=\textwidth]{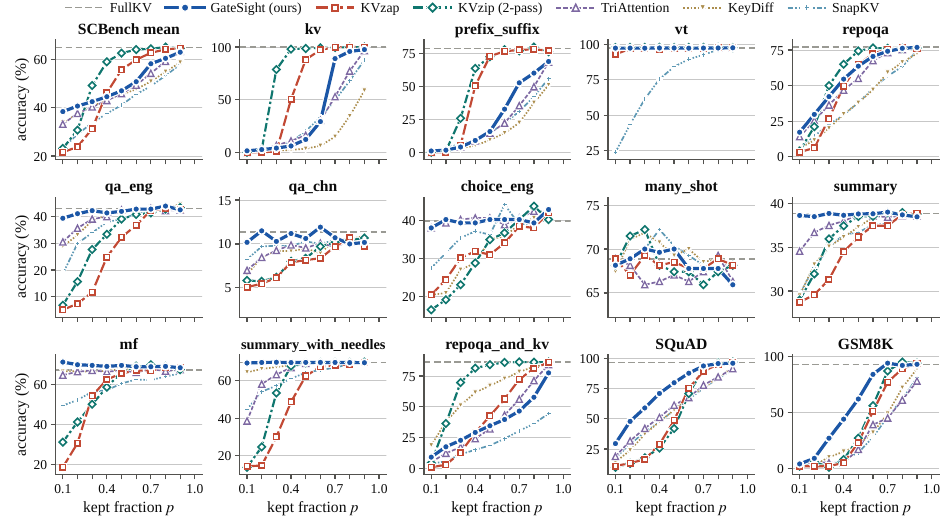}
\caption{Per-task accuracy (\%) vs.\ kept fraction $p$ on Qwen3.8-Flash-Next: SCBench, SQuAD v1.1, and GSM8K (dashed: FullKV).}
\label{fig:pertask_flashnext}
\end{minipage}
\end{figure*}

\begin{table*}[t]
\begin{minipage}{\textwidth}\centering\captionsetup{type=table}
{\footnotesize\renewcommand{\arraystretch}{\GSrowstretch}
\begin{tabular*}{\textwidth}{@{\extracolsep{\fill}}llcccccccccc}
\toprule
Model & Method & 0.9 & 0.8 & 0.7 & 0.6 & 0.5 & 0.4 & 0.3 & 0.2 & 0.1 & Mean \\
\midrule
\multirow{7}{*}{\begin{tabular}[c]{@{}l@{}}Qwen3.8-\\27B\end{tabular}} & FullKV & \multicolumn{10}{c}{97.5} \\
 & SnapKV & 75.3 & 69.5 & 64.8 & 61.2 & 56.5 & 52.9 & 49.0 & 43.0 & 35.4 & 56.4 \\
 & TriAttention & 97.3 & 96.6 & 94.1 & 90.4 & 85.6 & 76.6 & 63.1 & 48.9 & 36.4 & 76.6 \\
 & KeyDiff & 96.8 & 93.6 & 91.5 & 87.9 & 84.3 & 78.9 & 73.2 & 68.0 & 56.3 & 81.2 \\
 & KVzap & 97.5 & \textbf{97.6} & 97.6 & 97.3 & \textbf{97.4} & \textbf{96.7} & \textbf{93.9} & 80.6 & 38.8 & 88.6 \\
 & KVzip$^\dagger$ & 97.4 & \textbf{97.6} & \textbf{97.7} & \textbf{97.5} & \textbf{97.4} & 95.5 & 83.8 & 54.2 & 29.4 & 83.4 \\
\rowcolor{GSrow}\cellcolor{white} & \method & \textbf{97.6} & 97.5 & 97.4 & 97.4 & 97.0 & 95.2 & 89.9 & \textbf{82.6} & \textbf{76.5} & \textbf{92.3} \\
\midrule
\multirow{7}{*}{\begin{tabular}[c]{@{}l@{}}Qwen3.8-\\Flash-Next\end{tabular}} & FullKV & \multicolumn{10}{c}{96.8} \\
 & SnapKV & 76.2 & 68.4 & 62.6 & 57.8 & 54.0 & 50.3 & 44.7 & 37.2 & 27.1 & 53.1 \\
 & TriAttention & 92.1 & 82.0 & 72.5 & 63.3 & 54.2 & 47.4 & 41.2 & 36.9 & 26.7 & 57.4 \\
 & KeyDiff & 95.1 & 92.7 & 90.6 & 86.6 & 82.5 & 77.3 & \textbf{71.5} & \textbf{65.2} & \textbf{57.3} & \textbf{79.9} \\
 & KVzap & 95.9 & 94.2 & 91.2 & 84.2 & 69.7 & 49.4 & 32.7 & 23.4 & 21.1 & 62.4 \\
 & KVzip$^\dagger$ & \textbf{97.0} & \textbf{96.7} & \textbf{96.7} & \textbf{95.4} & \textbf{91.4} & \textbf{79.6} & 57.9 & 31.2 & 20.5 & 74.0 \\
\rowcolor{GSrow}\cellcolor{white} & \method & 96.5 & \textbf{96.7} & 95.8 & 90.1 & 83.0 & 76.2 & 69.0 & 60.2 & 47.5 & 79.4 \\
\bottomrule
\end{tabular*}
\par}
\caption{RULER-4096 13-task mean accuracy (\%) vs.\ kept fraction $p$ ($^\dagger$: two-pass).}
\label{tab:budgets_ruler}
\end{minipage}
\par\vspace{\GSgroupsep}
\begin{minipage}{\textwidth}\centering\captionsetup{type=figure}
\includegraphics[width=\textwidth]{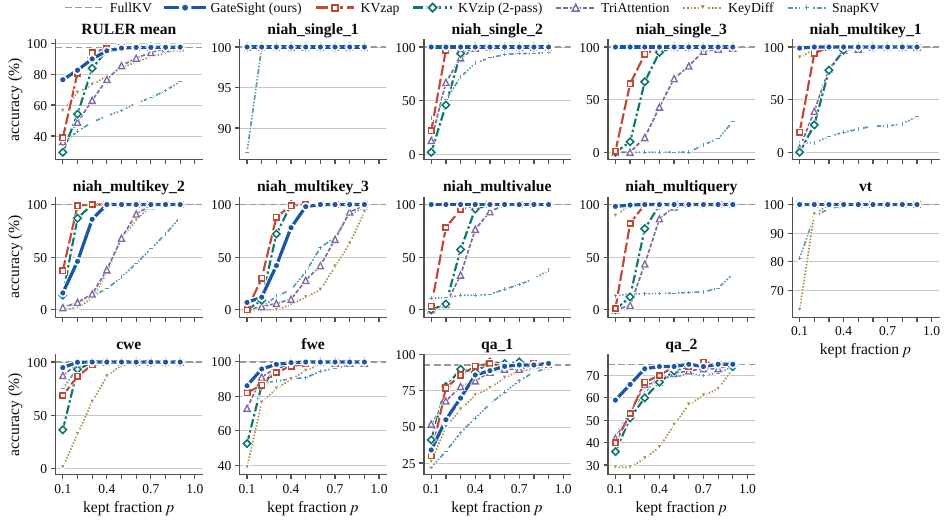}
\caption{Per-task RULER-4096 accuracy (\%) vs.\ kept fraction $p$ on Qwen3.8-27B (dashed: FullKV).}
\label{fig:ruler_pertask_27b}
\end{minipage}
\end{table*}

\begin{figure*}[t]
\begin{minipage}{\textwidth}\centering\captionsetup{type=figure}
\includegraphics[width=\textwidth]{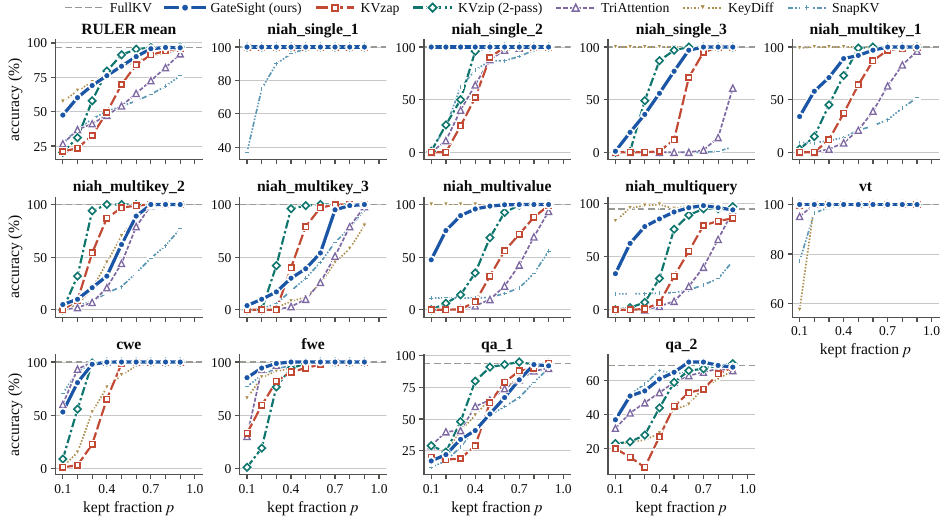}
\caption{Per-task RULER-4096 accuracy (\%) vs.\ kept fraction $p$ on Qwen3.8-Flash-Next (dashed: FullKV).}
\label{fig:ruler_pertask_flashnext}
\end{minipage}
\par\vspace{\GSgroupsep}
\begin{minipage}{\textwidth}\centering\captionsetup{type=figure}
\includegraphics[width=0.75\textwidth]{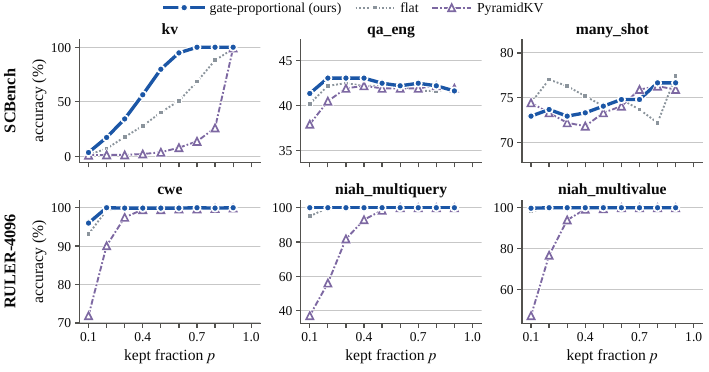}
\caption{Cross-layer allocation ablation per task on Qwen3.8-27B with fixed \method\ scores (accuracy vs.\ kept fraction $p$).}
\label{fig:ablation}
\end{minipage}
\end{figure*}

\paragraph{Corpus and Static Dynamics.}
The corpus of \S\ref{sec:obs} (benchmark names excluded) draws on nine Nemotron subsets: Nemotron-CC (high quality, synthetic, diverse QA, translated QA, math), synthetic code, and SFT (code, general, math). Static zero-input half-lives in \S\ref{sec:obs1} (analyzed on 240 documents, every tenth) are read directly from checkpoint parameters as $\ln 2 / \big(\exp(A_{\log})\,\mathrm{softplus}(\mathrm{dt}_{\mathrm{bias}})\big)$ (Figure~\ref{fig:obs1}). Pairwise head orderings by mean write $\beta\|v\|$ agree at Spearman $0.96$--$0.99$; at the $1{,}000$-token threshold, 173 of 2,304 heads are slow in $\ge 95\%$ of documents and 1,878 in $\le 5\%$. Retention $\exp(g)$ correlates with $\U$ at $-0.12$, mirroring forgetting $-g$, and Figure~\ref{fig:obs2example} shows a representative document trace.

\paragraph{Predictor Metrics and Signal Isolation.}
Predicted scores over $n$ context entries are evaluated against raw $\U$ via Spearman $\rho$, top-$k\%$ index overlap $\text{Ov@}k$ ($m_k=\max(4,\lfloor \frac{k}{100}n\rfloor)$), and utility-mass coverage $\text{Cov@}k$ (sum of $\U$ on predicted top-$m_k$ divided by true top-$m_k$ sum), averaged across KV heads, documents, and 16 attention layers. Untrained signals run on all 2,428 documents; learned predictors use 9-subset LOCO cross-validation, while the single-head oracle picks the (\gdn\ layer, head) cell maximizing in-sample Ov@10 (Table~\ref{tab:obs3full}; Figures~\ref{fig:obs2layers}--\ref{fig:obs3layers}). The 864-feature MLP beats ridge on Ov@10 in 8/9 subsets (widest on SFT code: $44.1\%$ vs.\ $35.5\%$). Table~\ref{tab:signal_mlp} isolates input signals by training matched width-64 BCE MLPs under the 240-document LOCO protocol on probe-free intra-layer key inputs, namely TriAttention (calibrated excluding held-out documents) and KeyDiff, without positional features.

\paragraph{Length and Task Transfer.}
Table~\ref{tab:transfer} evaluates the unchanged 1K-trained estimator on Nemotron documents truncated to 2K--8K tokens and on RULER-8K contexts (12 items each of \texttt{niah\_single\_1}, \texttt{niah\_multikey\_3}, \texttt{qa\_1}, \texttt{cwe}, \texttt{vt}; 7.6K--11.1K tokens), with the largest gains over slow-head $\beta\|v\|$ on \texttt{cwe} ($36.1\%$ vs.\ $26.3\%$ Ov@10), \texttt{qa\_1} ($31.6\%$ vs.\ $23.4\%$), and \texttt{niah\_multikey\_3} ($37.7\%$ vs.\ $30.3\%$). Longer contexts are evaluated end to end on the downstream benchmarks of \S\ref{sec:exp}.

\section{Output-Gate Ablation}
\label{app:gate}

Supplementing \S\ref{sec:allocation} ($w_\ell$ is calibrated on $2{,}048$ wiki prefill tokens on 27B and aggregated directly from prefill gates recorded over the $2{,}428$ Nemotron training documents on Flash-Next; document-level $w_\ell$ rankings match the global profile at median $\rho = 0.969$ and per-prompt recomputation shifts $\approx\!1\%$ of block budgets, making $w_\ell$ insensitive to the calibration corpus), three diagnostics on Qwen3.8-27B validate gate-proportional per-layer budgets:
(1)~Cross-head hierarchy: zeroing the 48 lowest-gate query heads (of 384) incurs $2\text{--}4\times$ smaller next-token accuracy loss on wiki, code, and math text than zeroing 48 random heads, while within-head gate rescaling leaves eviction accuracy unchanged.
(2)~Per-layer vs.\ per-head splits: on five-task RULER-8K at $p\in\{0.05,0.1,0.2,0.5\}$, per-layer gate budgets achieve $(65.7, 71.0, 80.0, 95.0)\%$ vs.\ $(65.9, 71.0, 74.0, 92.0)\%$ for per-head splits and $(63.1, 68.9, 74.0, 87.0)\%$ for flat budgets.
(3)~Exponent sweep: on solvable RULER-8K tasks at $p=0.05$, unit proportionality $K_\ell \propto w_\ell^{1.0}$ ($61.5\%$) outperforms $w_\ell^{0.5}$ ($59.2\%$), $w_\ell^{1.5}$ ($60.8\%$), and flat ($53.8\%$).

\section{Per-Task and Per-Budget Results}
\label{app:pertask}

Tables~\ref{tab:budgets_scbench}--\ref{tab:budgets_gsm8k} report nine-budget means and Figures~\ref{fig:pertask}--\ref{fig:pertask_flashnext} plot each task. Per-task curves show that KVzip and KVzap gain their mid-budget verbatim edge primarily on unstructured string and code copying (\texttt{kv}, \texttt{prefix\_suffix}, \texttt{repoqa}, \texttt{repoqa\_and\_kv}), whereas on multi-hop variable tracking (\texttt{vt}) \method\ leads KVzap across $p\le 0.5$ by up to $+30.7$ points ($89.6\%$ vs.\ $78.2\%$ mean on 27B), with Flash-Next semantic gains reaching $+52.3$ points on \texttt{mf} and $+34.2$ on \texttt{qa\_eng} at $p=0.1$.

\section{RULER-4096 Diagnostic Results}
\label{app:ruler}

Table~\ref{tab:budgets_ruler} and Figures~\ref{fig:ruler_pertask_27b}--\ref{fig:ruler_pertask_flashnext} report all 13 synthetic RULER-4096 tasks. On Qwen3.8-27B at $p=0.1$, \method\ holds $98.2\%$--$100.0\%$ across seven needle and variable-tracking tasks, on five of which KVzap and KVzip drop to $0\%$--$22\%$ under their native configurations. On Qwen3.8-Flash-Next, \method\ leads KeyDiff at all $p\ge 0.5$; the lead of KeyDiff at $p\le 0.4$ (nine-budget mean $79.9\%$ vs.\ $79.4\%$) stems primarily from four repetitive-essay needle tasks where rare needle keys stand out from the mean key.

\section{Ablations and Diagnostic Analyses}
\label{app:ablation}

\paragraph{Cross-Layer Allocation.}
Forking equal-budget caches from identical \method\ scores (Table~\ref{tab:ablation}, Figure~\ref{fig:ablation}), bounded gate-proportional allocation exceeds flat budgets in mean accuracy on all seven suites and PyramidKV~\citep{cai2024pyramidkv} (schedule hyperparameter $20$, floor of 4 entries) on both suites where it is run, with the largest nine-budget gains on \texttt{kv} ($+20.7$ points over flat and $+47.8$ over PyramidKV).

\begin{table}[t]
\centering
\footnotesize
\setlength{\tabcolsep}{2.5pt}
\begin{tabular*}{\linewidth}{@{\extracolsep{\fill}}lcccc}
\toprule
Allocation & $p=0.7$ & $0.4$ & $0.1$ & Mean \\
\midrule
\multicolumn{5}{@{}l}{\textit{SCBench (27B: \textup{\texttt{kv}}, \textup{\texttt{qa\_eng}}, \textup{\texttt{many\_shot}})}} \\[1pt]
\rowcolor{GSrow}Gate-proportional (ours) & \textbf{72.4} & \textbf{57.6} & \textbf{39.2} & \textbf{60.6} \\
Flat (uniform) & 61.2 & 48.4 & 38.6 & 53.7 \\
PyramidKV & 43.8 & 38.7 & 37.8 & 44.3 \\
\midrule
\multicolumn{5}{@{}l}{\textit{RULER-4096 (27B: \textup{\texttt{cwe}}, \textup{\texttt{multiquery}}, \textup{\texttt{multivalue}})}} \\[1pt]
\rowcolor{GSrow}Gate-proportional (ours) & \textbf{100.0} & \textbf{100.0} & \textbf{98.6} & \textbf{99.8} \\
Flat (uniform) & \textbf{100.0} & \textbf{100.0} & 95.5 & 99.4 \\
PyramidKV & 99.9 & 97.2 & 52.1 & 90.4 \\
\midrule
\multicolumn{5}{@{}l}{\textit{SQuAD v1.1 (Qwen3.8-27B)}} \\[1pt]
\rowcolor{GSrow}Gate-proportional (ours) & \textbf{97.5} & \textbf{93.9} & \textbf{54.4} & \textbf{88.9} \\
Flat (uniform) & 97.0 & 89.6 & 46.2 & 85.9 \\
\midrule
\multicolumn{5}{@{}l}{\textit{SQuAD v1.1 (Qwen3.8-Flash-Next)}} \\[1pt]
\rowcolor{GSrow}Gate-proportional (ours) & \textbf{93.8} & \textbf{71.0} & \textbf{29.5} & \textbf{73.4} \\
Flat (uniform) & 87.8 & 66.2 & 27.0 & 69.5 \\
\midrule
\multicolumn{5}{@{}l}{\textit{GSM8K (Qwen3.8-27B)}} \\[1pt]
\rowcolor{GSrow}Gate-proportional (ours) & \textbf{92.0} & 90.0 & \textbf{43.0} & \textbf{82.3} \\
Flat (uniform) & 90.0 & \textbf{91.0} & 30.0 & 78.2 \\
\midrule
\multicolumn{5}{@{}l}{\textit{GSM8K (Qwen3.8-Flash-Next)}} \\[1pt]
\rowcolor{GSrow}Gate-proportional (ours) & \textbf{94.0} & \textbf{44.0} & 4.0 & \textbf{56.6} \\
Flat (uniform) & 76.0 & 36.0 & \textbf{5.0} & 49.9 \\
\midrule
\textit{AIME24/25 (27B)} & \textit{4,096} & \textit{3,072} & \textit{2,048} & \textit{Mean} \\[1pt]
\rowcolor{GSrow}Gate-proportional (ours) & \textbf{93.3} & \textbf{91.7} & \textbf{86.7} & \textbf{90.6} \\
Flat (uniform) & 90.0 & 81.7 & 81.7 & 84.4 \\
\bottomrule
\end{tabular*}

\caption{Cross-layer allocation ablation with fixed \method\ scores (accuracy, \%). Columns: kept fraction $p$ or AIME cache cap; Mean: over all nine $p$ or the three caps.}
\label{tab:ablation}
\end{table}

\paragraph{Supervision Target vs.\ Allocation on Verbatim Tasks.}
To diagnose the mid-budget verbatim gap (\S\ref{sec:prefill}), we evaluate Qwen3.8-Flash-Next across the first 50 contexts of \texttt{kv} (250 queries), \texttt{prefix\_suffix} (250), and \texttt{repoqa\_and\_kv} (400) at $p=0.4$ (Table~\ref{tab:supervision}), training identical \gdn\ MLPs under flat budgets on two disjoint 1,090-document Nemotron cohorts (A/B, shared 248-doc validation) against self-continuation utility (\texttt{SU}) or teacher-forced reconstruction attention (\texttt{RA}). In both cohorts, \texttt{RA} raises mixed code+key retrieval (\texttt{repoqa\_and\_kv}) and lowers random-UUID lookup (\texttt{kv}, \texttt{prefix\_suffix}). With the deployed scorer fixed, gate-proportional budgets beat flat budgets on all three tasks, while two-pass KVzip reaches $98.0\%$, $78.1\%$, and $82.9\%$ under identical 4-token QSA repacking, locating the verbatim deficit in entry scoring rather than QSA repacking or gate allocation.

\begin{table}[t]
\centering
\footnotesize
\setlength{\tabcolsep}{3.5pt}
\begin{tabular*}{\linewidth}{@{\extracolsep{\fill}}lcccccc}
\toprule
 & \multicolumn{2}{c}{Cohort A} & \multicolumn{2}{c}{Cohort B} & \multicolumn{2}{c}{Allocation} \\
\cmidrule(lr){2-3}\cmidrule(lr){4-5}\cmidrule(lr){6-7}
SCBench task & SU & RA & SU & RA & Flat & Gate \\
\midrule
\texttt{kv} & \textbf{21.2} & 2.0 & \textbf{28.8} & 0.8 & 14.4 & \textbf{20.8} \\
\texttt{prefix\_suffix} & \textbf{9.6} & 2.4 & \textbf{4.0} & 0.8 & 4.8 & \textbf{7.2} \\
\texttt{repoqa\_and\_kv} & 33.8 & \textbf{57.5} & 30.8 & \textbf{47.3} & 29.0 & \textbf{31.3} \\
\bottomrule
\end{tabular*}

\caption{Supervision target (SU: self-continuation utility; RA: reconstruction attention) and allocation (\texttt{flat} vs.\ gate-proportional) on Qwen3.8-Flash-Next at $p=0.4$ (accuracy, \%).}
\label{tab:supervision}
\end{table}

\begin{table}[t]
\centering
\footnotesize
\setlength{\tabcolsep}{1.8pt}
\begin{tabular*}{\linewidth}{@{\extracolsep{\fill}}ccccccc}
\toprule
$W$ & Inputs & Params & $\rho$ & Ov@5 & Ov@10 & Cov@10 \\
\midrule
3 & 432 & 1.8M & $+0.515$ & 29.3 & 37.7 & 54.1 \\
\rowcolor{GSrow}\textbf{6} & 864 & 3.4M & $+0.523$ & 29.8 & 38.1 & 54.6 \\
9 & 1,296 & 5.0M & $+0.524$ & 29.9 & 38.2 & 54.7 \\
12 & 1,728 & 6.4M & $+0.524$ & 29.9 & 38.1 & 54.7 \\
\bottomrule
\end{tabular*}

\caption{Window sweep on Qwen3.8-27B under identical hyperparameters ($W$: preceding \gdn\ layers read; LOCO means over 16 attention layers; Ov/Cov in \%; shaded: deployed).}
\label{tab:window}
\end{table}

\section{Scoring Latency and Throughput}
\label{app:cost}

Tables~\ref{tab:cost}--\ref{tab:cost_flashnext} isolate post-prefill scorer latency across 4K--256K tokens (plain 256K prefill: $46.26$~s on 27B and $13.94$~s on Flash-Next); unfused Python prefill hooks on 27B separately add $3.67$~s ($7.9\%$) for \gdn\ scalars vs.\ $4.95$~s ($10.7\%$) for SnapKV/KVzap tensors, both eliminable via runner/kernel fusion. Table~\ref{tab:serving} measures decoding throughput ($\text{Peak }B/\text{TPOT}$) over $128$ generated tokens after $260{,}096$-token prefill (shared initial \gdn\ snapshot, independent forked KV pages). Beyond the batch-capacity gains in \S\ref{sec:eff}, single-request TPOT on 27B drops at $p=0.1$ from $112.3\to 67.2$~ms/tok (eager) and $111.6\to 35.1$~ms/tok (CUDA graphs), whereas the constant-window 2,048-token QSA of Flash-Next realizes its throughput gain via batch expansion.

\begin{table}[t]
\centering
\footnotesize
\setlength{\tabcolsep}{4pt}
\begin{tabular*}{\linewidth}{@{\extracolsep{\fill}}lrrrr}
\toprule
\multirow{2}{*}{Method} & \multicolumn{4}{c}{Context length (tokens)} \\
\cmidrule(lr){2-5}
 & 4K & 16K & 64K & 256K \\
\midrule
\rowcolor{GSrow}\method & 0.004 & 0.018 & 0.067 & 0.254 \\
TriAttention & 0.013 & 0.016 & 0.034 & 0.108 \\
KeyDiff & 0.003 & 0.007 & 0.023 & 0.084 \\
SnapKV & 0.006 & 0.016 & 0.058 & 0.285 \\
KVzap & 0.016 & 0.065 & 0.268 & 1.073 \\
KVzip$^\dagger$ & 0.741 & 2.769 & 13.487 & 99.222 \\
\bottomrule
\end{tabular*}

\caption{vLLM scoring latency (s) on Qwen3.8-27B ($1\times$H200, median of 3 runs; $^\dagger$: two-pass).}
\label{tab:cost}
\end{table}

\begin{table}[t]
\centering
\footnotesize
\setlength{\tabcolsep}{4pt}
\begin{tabular*}{\linewidth}{@{\extracolsep{\fill}}lrrrr}
\toprule
\multirow{2}{*}{Method} & \multicolumn{4}{c}{Context length (tokens)} \\
\cmidrule(lr){2-5}
 & 4K & 16K & 64K & 256K \\
\midrule
\rowcolor{GSrow}\method & 0.004 & 0.013 & 0.049 & 0.200 \\
TriAttention & 0.010 & 0.011 & 0.012 & 0.037 \\
KeyDiff & 0.004 & 0.004 & 0.006 & 0.030 \\
SnapKV & 0.006 & 0.006 & 0.011 & 0.053 \\
KVzap & 0.005 & 0.015 & 0.059 & 0.247 \\
KVzip$^\dagger$ & 0.557 & 1.882 & 8.186 & 52.811 \\
\bottomrule
\end{tabular*}

\caption{vLLM scoring latency (s) on Qwen3.8-Flash-Next ($2\times$H200, TP=2, median of 3 runs; $^\dagger$: two-pass).}
\label{tab:cost_flashnext}
\end{table}

\begin{table}[t]
\centering
\footnotesize
\setlength{\tabcolsep}{2.6pt}
\begin{tabular*}{\linewidth}{@{\extracolsep{\fill}}lccccc}
\toprule
Model / mode & $p$ & $B_{\max}$ & Peak $B$ & TPOT & tok/s \\
\midrule
27B / eager & 1.0 & 4 & 4 & 117.0 & 34.2 \\
27B / eager & 0.7 & 6 & 6 & 102.8 & 58.4 \\
\rowcolor{GSrow}27B / eager & 0.1 & 41 & 41 & 117.0 & \textbf{350.4} \\
\midrule
27B / graphs & 1.0 & 4 & 4 & 116.2 & 34.4 \\
27B / graphs & 0.7 & 6 & 6 & 93.9 & 63.9 \\
\rowcolor{GSrow}27B / graphs & 0.1 & 41 & 32 & 76.5 & \textbf{418.5} \\
\midrule
Flash / eager & 1.0 & 9 & 9 & 121.5 & 74.1 \\
Flash / eager & 0.9 & 10 & 10 & 126.4 & 79.1 \\
\rowcolor{GSrow}Flash / eager & 0.1 & 86 & 86 & 221.6 & \textbf{388.2} \\
\bottomrule
\end{tabular*}

\caption{260K H200 serving: largest batch $B_{\max}$, batch at peak throughput (Peak~$B$), TPOT (ms/tok), and peak decoding throughput (tok/s) over 128 generated tokens (shaded: $p=0.1$).}
\label{tab:serving}
\end{table}

\section{Implementation Details}
\label{app:details}

\paragraph{Training, Allocation, and Compute.}
Supplementing \S\ref{sec:estimator}--\S\ref{sec:allocation}, MLPs use AdamW weight decay $10^{-2}$, gradient clip $1.0$, and seed $0$ (max-normalizing 1K-doc $\U_i^{(\ell,h)}$ onto $[0,1]$ stabilizes heavy-tailed BCE training while leaving inference top-$K_\ell$ rankings scale-invariant); Eq.~\ref{eq:alloc} uses largest-remainder rounding to sum to $B$. Labeling/training use \texttt{transformers} 5.15.1 (one RTX PRO 6000 Blackwell 96~GB for 27B labeling; one H200 141~GB for Flash-Next training and 27B serving; $2\times$H200 TP=2 for Flash-Next SCBench on vLLM 0.28 / PyTorch 2.13).

\paragraph{Baseline Configurations.}
Under vLLM's per-layer block tables: SnapKV uses a 32-token tail window with kernel-7 max pooling and flat budgets; TriAttention calibrates pre-RoPE query statistics on 32K Nemotron tokens (rotary factor $0.25$, 64/256 rotated dims, GQA max aggregation, flat budgets); KeyDiff uses cosine dissimilarity to the unnormalized post-RoPE head mean under flat budgets; KVzap~\citep{jegou2026kvzap} trains two-layer GELU MLPs on post-input-norm hidden states ($5120\to640\to4$ on 27B; $2560\to512\to2$ on Flash-Next; 15 epochs, AdamW $\text{lr}=5\times10^{-3}$ cosine decay, MSE on $\log\max(\text{KVzip+},10^{-6})$ with $\text{KVzip+}=\max_{j,h}a_{j,i}^{(h)}\|W_O^{(h)}v_i\|_2/\|h_j\|_2$) and allocates per-layer totals via global top-$B$ sorting (the fixed-budget counterpart of its global score threshold) split evenly across KV heads; KVzip accumulates two-pass reconstruction attention under the same global top-$B$ rule; StreamingLLM (decoding) keeps 4 sink plus $\text{budget}-4$ recent tokens.

\paragraph{Per-Head Ragged Paging Overhead.}
Replacing per-layer block tables on Qwen3.8-27B with a custom ragged per-head block pool increases decoding TPOT by $+9.4\%$--$11.7\%$ at 4K (42 configs) and by $+10.6\%$, $+22.7\%$, and $+29.5\%$ at $p=0.1, 0.3, 0.5$ on 260K contexts (single request: $69.7\to 77.1$, $71.7\to 88.0$, $77.8\to 100.7$~ms/tok), while per-layer \method\ still beats ragged per-head KVzip on RULER-4096 by $+32.5$, $+11.4$, and $+1.5$ points at $p=0.1, 0.2, 0.3$.

\section{Qwen3.8-Flash-Next Details}
\label{app:flashnext}

Qwen3.8-Flash-Next (\texttt{Qwen4\allowbreak Exp\allowbreak For\allowbreak Conditional\allowbreak Generation}, vLLM commit \texttt{c5d840ff6} with CPU PLE offload) uses 4-token QSA sparse attention in its 12 attention layers (24 query, 2 KV heads) alongside a short-convolution state (FP8 routed experts, BF16 elsewhere, 262K max context). Its estimators (\S\ref{sec:estimator}) are supervised via Eq.~\ref{eq:utility} over QSA-selected token sets using dequantized FP8 $W_O$ (relative $L_2$ check $\le 5\%$). Because both KV heads in a QSA layer share one 4-token block list, the block pooling of \S\ref{sec:deploy} is applied to all methods. On RULER-4096, gate-proportional budgets outperform flat budgets at every kept fraction by $0.3$--$2.7$ points.

\end{document}